\documentclass[10pt,twocolumn,letterpaper,table,dvipsnames]{article}

\usepackage{verbatim}
\usepackage{bm}
\usepackage{bbm}
\usepackage{multirow}
\usepackage{titletoc}
\usepackage{array}
\usepackage{pifont}
\usepackage{nccmath}
\usepackage{titletoc}
\usepackage{subcaption}
\usepackage[table]{xcolor}
\usepackage{physics}
\usepackage{amsmath}
\usepackage{tikz}
\usepackage{mathdots}
\usepackage{yhmath}
\usepackage{cancel}
\usepackage{color}
\usepackage{siunitx}
\usepackage{array}
\usepackage{multirow}
\usepackage{amssymb}
\usepackage{gensymb}
\usepackage{tabularx}
\usepackage{extarrows}
\usepackage{booktabs}
\usetikzlibrary{fadings}
\usetikzlibrary{patterns}
\usetikzlibrary{shadows.blur}
\usetikzlibrary{shapes}

\usepackage[linesnumbered,ruled,vlined]{algorithm2e}
\usepackage{chngcntr}
\usepackage{tocloft}
\usepackage[titletoc,toc,title]{appendix}

\def \vc {\bm{v}_c}
\def \vp {\bm{v}_p}
\def \vf {\bm{v}_f}

\def \zp {\bm{z}_p}
\def \zf {\bm{z}_f}
\def \zcs {\bm{z}_c}

\def \zcsp {\bm{z}_c^p}
\def \zcsf {\bm{z}_c^f}

\def \lpt {\mathcal{L}_{pt}}
\def \lft {\mathcal{L}_{ft}}
\def \lpf {\mathcal{L}_{pf}}

\definecolor{soft_red}{RGB}{240, 110, 110}
\definecolor{soft_blue}{RGB}{100, 170, 220}
\definecolor{define_red}{RGB}{255, 20, 20}
\definecolor{define_green}{RGB}{55, 100, 0}

\usepackage[pagenumbers]{cvpr} % To force page numbers, e.g. for an arXiv version

\definecolor{cvprblue}{rgb}{0.21,0.49,0.74}

\usepackage[pagebackref,breaklinks,colorlinks,anchorcolor=cvprblue, citecolor=cvprblue,linkcolor=cvprblue]{hyperref}

\def\paperID{13020} % *** Enter the Paper ID here
\def\confName{CVPR}
\def\confYear{2025}

\title{When the Future Becomes the Past: Taming Temporal Correspondence \\ for Self-supervised Video Representation Learning}

\author{
	Yang Liu\textsuperscript{1}\hspace{2em} Qianqian Xu\textsuperscript{2,}\thanks{Corresponding authors}\hspace{2em} Peisong Wen\textsuperscript{1,2}\hspace{2em} Siran Dai\textsuperscript{3,4}\hspace{2em} Qingming Huang\textsuperscript{1,2,*} \\
	{\textsuperscript{1}School of Computer Science and Technology, University of Chinese Academy of Sciences} \\
	{\textsuperscript{2}Institute of Computing Technology, Chinese Academy of Sciences} \\
    {\textsuperscript{3}Institute of Information Engineering, Chinese Academy of Sciences} \\
    {\textsuperscript{4}School of Cyber Security, University of Chinese Academy of Sciences} \\
	{\tt\small liuyang232@mails.ucas.ac.cn\hspace{2em} \{xuqianqian, wenpeisong20z\}@ict.ac.cn} \\ 
    {\tt\small daisiran@iie.ac.cn\hspace{2em} qmhuang@ucas.ac.cn }
}

\begin{document}
\maketitle
\begin{abstract}

The past decade has witnessed notable achievements in self-supervised learning for video tasks.
Recent efforts typically adopt the Masked Video Modeling (MVM) paradigm, leading to significant progress on multiple video tasks.
However, two critical challenges remain:
\textbf{1)} Without human annotations, the random temporal sampling introduces uncertainty, increasing the difficulty of model training.
\textbf{2)} Previous MVM methods primarily recover the masked patches in the pixel space, leading to insufficient information compression for downstream tasks.
To address these challenges jointly, we propose a self-supervised framework that leverages \textbf{T}emporal \textbf{Co}rrespondence for video \textbf{Re}presentation learning (\textbf{T-CoRe}). 
For challenge \textbf{1)}, we propose a sandwich sampling strategy that selects two auxiliary frames to reduce reconstruction uncertainty in a two-side-squeezing manner.
Addressing challenge \textbf{2)}, we introduce an auxiliary branch into a self-distillation architecture to restore representations in the latent space, generating high-level semantic representations enriched with temporal information.
Experiments of T-CoRe consistently present superior performance across several downstream tasks, demonstrating its effectiveness for video representation learning.
The code is available at \href{https://github.com/yafeng19/T-CORE}{https://github.com/yafeng19/T-CORE}.

% The past decade has witnessed notable achievements in self-supervised learning for video tasks. Recent efforts typically adopt the Masked Video Modeling (MVM) paradigm, leading to significant progress on multiple video tasks. However, two critical challenges remain: 1) Without human annotations, the random temporal sampling introduces uncertainty, increasing the difficulty of model training. 2) Previous MVM methods primarily recover the masked patches in the pixel space, leading to insufficient information compression for downstream tasks. To address these challenges jointly, we propose a self-supervised framework that leverages Temporal Correspondence for video Representation learning (T-CoRe). For challenge 1), we propose a sandwich sampling strategy that selects two auxiliary frames to reduce reconstruction uncertainty in a two-side-squeezing manner. Addressing challenge 2), we introduce an auxiliary branch into a self-distillation architecture to restore representations in the latent space, generating high-level semantic representations enriched with temporal information. Experiments of T-CoRe consistently present superior performance across several downstream tasks, demonstrating its effectiveness for video representation learning. The code is available at https://github.com/yafeng19/T-CORE.

\end{abstract}
    
\section{Introduction}
\label{sec:introduction}

\begin{figure}
  \centering
    \begin{subfigure}{0.49\linewidth}
    \includegraphics[width=1.0\linewidth]{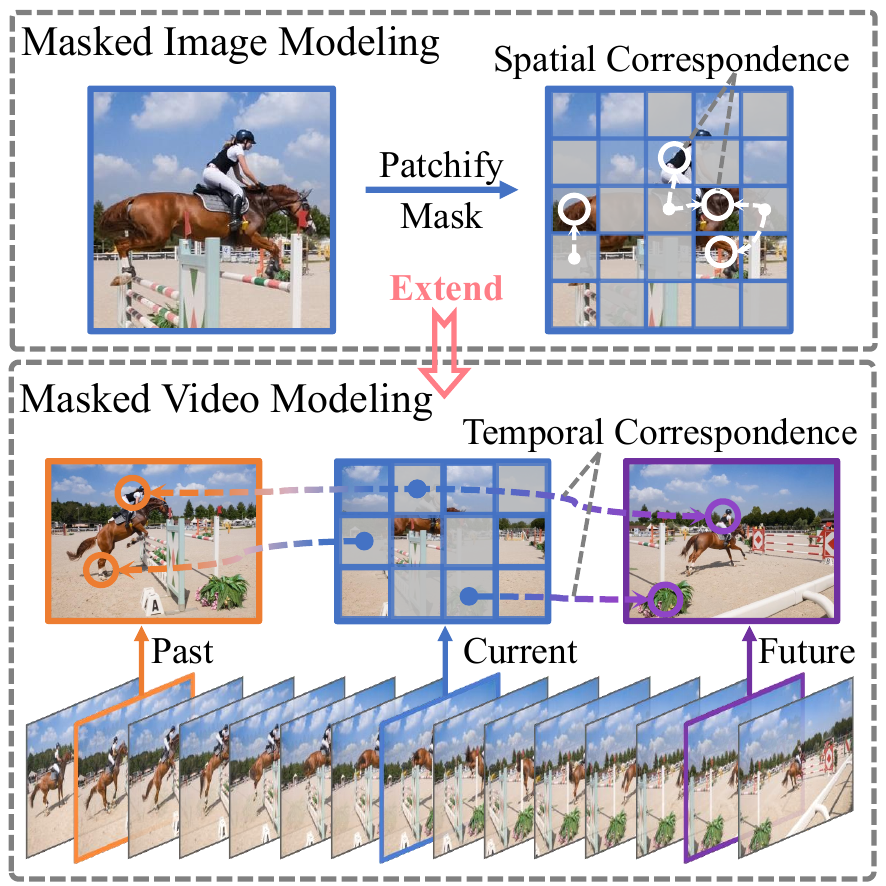}
    \caption{Extension from MIM to MVM.}
    \label{fig:intro}
    \end{subfigure}
  \hfill
  \begin{subfigure}{0.49\linewidth}
    \includegraphics[width=1.0\linewidth]{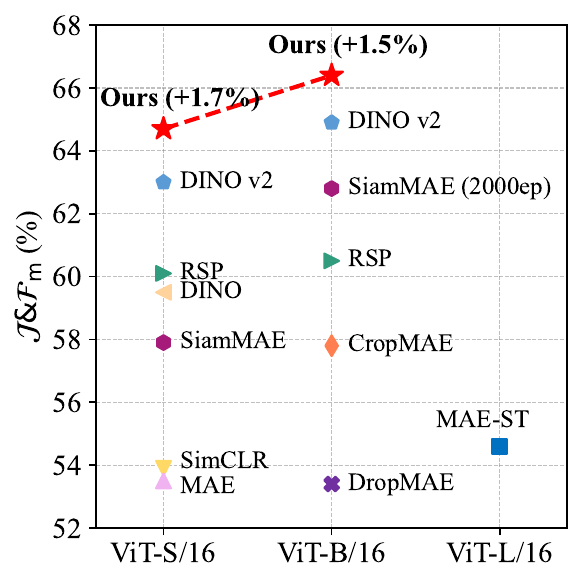}
    \caption{$\mathcal{J}\&\mathcal{F}_{\mathrm{m}}$ on DAVIS-2017.}
    \label{fig:scatter_DAVIS}
  \end{subfigure}
  \caption{(a) MIM restores the masked image patches with the assistance of spatially adjacent patches, while MVM restores the masked frame patches by establishing correspondence from temporally adjacent frames. (b) Our framework outperforms state-of-the-art methods on multiple video downstream tasks such as video object segmentation on DAVIS-2017~\cite{DAVIS17}.}
  \vspace{-12pt}
  \label{fig:intro_total}
\end{figure}

Self-supervised representation learning has achieved remarkable progress by effectively exploiting rich visual information without human annotations~\cite{MoCov1,BYOL,DINO,MAE,BEiTv1,iBOT}. 
One of the most promising self-supervised approaches is \textit{\textbf{M}asked \textbf{I}mage \textbf{M}odeling (\textbf{MIM})}~\cite{MAE,BEiTv1,BEiTv2}, in which the model learns representations by reconstructing the masked regions of an image using an encoder-decoder architecture.

The success of MIM has inspired its extension to the video domain, referred to \textit{\textbf{M}asked \textbf{V}ideo \textbf{M}odeling (\textbf{MVM})}. A key difference in video data is the additional temporal dimension. 
As shown in ~\cref{fig:intro}, MIM restores the masked patches by using spatially adjacent patches. In a similar vein, MVM extends this idea into the temporal dimension, reconstructing masked frame patches by establishing correspondences with temporally adjacent frames. 
This extension raises two critical questions: \textbf{1) Which} frames should be selected by a \textit{sampling strategy} to establish temporal correspondence? \textbf{2) How} can we effectively \textit{reconstruct representations} of videos to capture inter-frame correlations?

\begin{figure}
  \centering
    \includegraphics[width=0.96\linewidth]{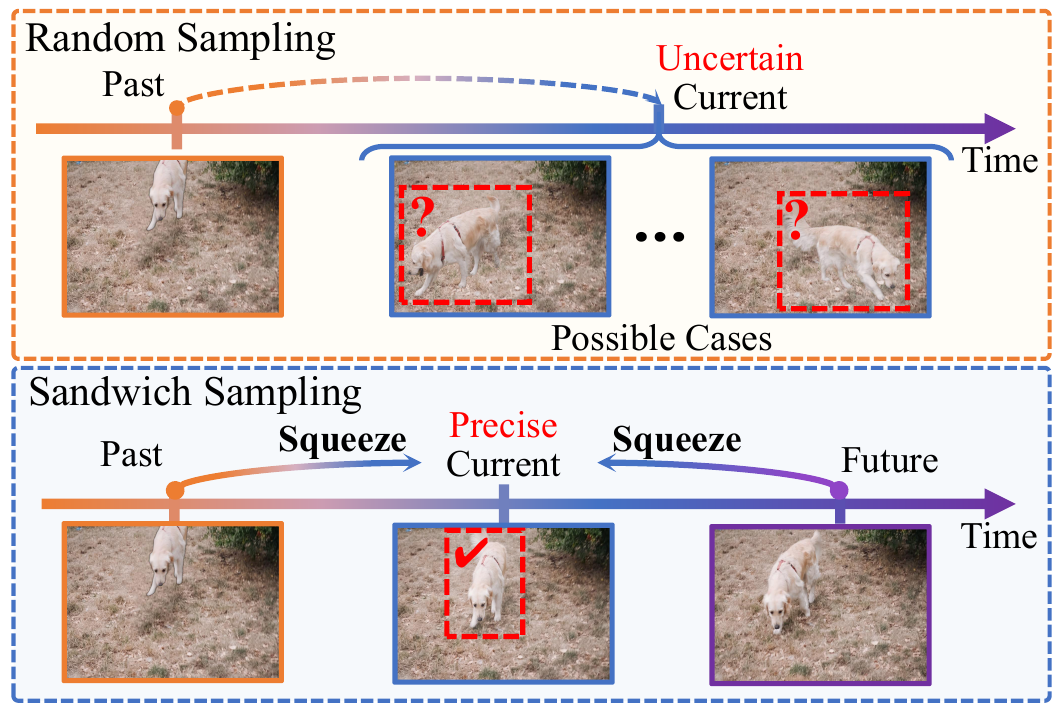}
    \caption{Top: inferring the object in the current frame from a random past frame suffers from high uncertainty. Bottom: our sandwich sampling reduces the uncertainty in a two-side-squeezing manner, such that the current object can be precisely located.}
    \vspace{-12pt}
    \label{fig:motivation}
\end{figure}

Early efforts primarily adopt a dense sampling strategy, where all frames are directly fed into a spatiotemporal encoder~\cite{MAE_ST,VideoMAE,VideoMAEv2,BEVT,DropMAE}. These methods implicitly embed the sampled frames by forcing the model to restore the masked patches.
Without explicit mechanisms to guide the process, these approaches demand large models and extensive datasets to capture temporal structures in videos, resulting in high computational costs.
More recent methods have shifted to random sampling techniques, where masked frame patches are restored using another frame as prior information based on a conditional decoder~\cite{SiamMAE,CropMAE,STP,RSP}. However, as illustrated in \cref{fig:motivation}, a single conditional frame can generate multiple potential predictions, introducing uncertainty in the reconstruction of representations and confusing the model in complex dynamic scenes ~\cite{RSP,uncertainty_1,uncertainty_2}.

Regarding the reconstruction approach, previous MVM methods mainly focus on recovering masked content in pixel space. Since video content is highly redundant, restoring all masked pixels forces the model to retain excessive low-level details, which hinders learning high-level semantics. Thus, the lack of information compression~\cite{compression_1,compression_2,compression_3} results in suboptimal performance on downstream tasks, as the model tends to overfit to superficial features.

To address issue \textbf{1)}, we propose a \textit{sandwich sampling strategy} for establishing temporal correspondence. Specifically, as depicted in \cref{fig:motivation}, for a given masked frame, we sample two auxiliary frames: one from the past and one from the future. Afterward, the encoder restores representations of the masked patches with the assistance of two auxiliary frames. In this way, the uncertainty of the reconstruction can be reduced in a two-side-squeezing manner.

In search of a solution to the issue \textbf{2)}, we propose reconstructing the masked patches in latent space. On top of the two-branch self-distillation architecture of DINO~\cite{DINO,DINOv2}, we introduce an auxiliary branch equipped with our designed \textit{\textbf{P}atch \textbf{M}atching \textbf{M}odule (\textbf{PMM})}.
In complement to learning spatial information from unmasked patches within the current frame, this branch leverages the auxiliary frames to provide temporal references for representation reconstruction. This helps the model reduce its reliance on redundant low-level visual features and, instead, focus on learning more robust, high-level semantic representations.

By integrating the above techniques, we propose a self-supervised framework, called \textit{\textbf{T}emporal \textbf{Co}rrespondence for video \textbf{Re}presentation learning (\textbf{T-CoRe})}. As presented in ~\cref{fig:scatter_DAVIS}, our T-CoRe consistently outperforms state-of-the-art methods under the same backbone.

In summary, the contributions are three-fold:
\begin{itemize}
  \item We propose T-CoRe, a self-supervised framework for video representation learning that incorporates temporal correspondence to guide the reconstruction of masked patches in latent space.
  \item Within T-CoRe, we introduce a sandwich sampling strategy to reduce reconstruction uncertainty and integrate an auxiliary branch with the PMM into a self-distillation architecture, enabling the model to learn high-level semantic representations with temporal information.
  \item Experimental results show that T-CoRe consistently outperforms previous methods across three video downstream tasks, demonstrating its effectiveness in video representation learning that facilitates video understanding.
\end{itemize}

\section{Related Work}
\label{sec:related_work}

\begin{figure*}
\vspace{-10pt}
  \centering
    \includegraphics[width=0.98\linewidth]{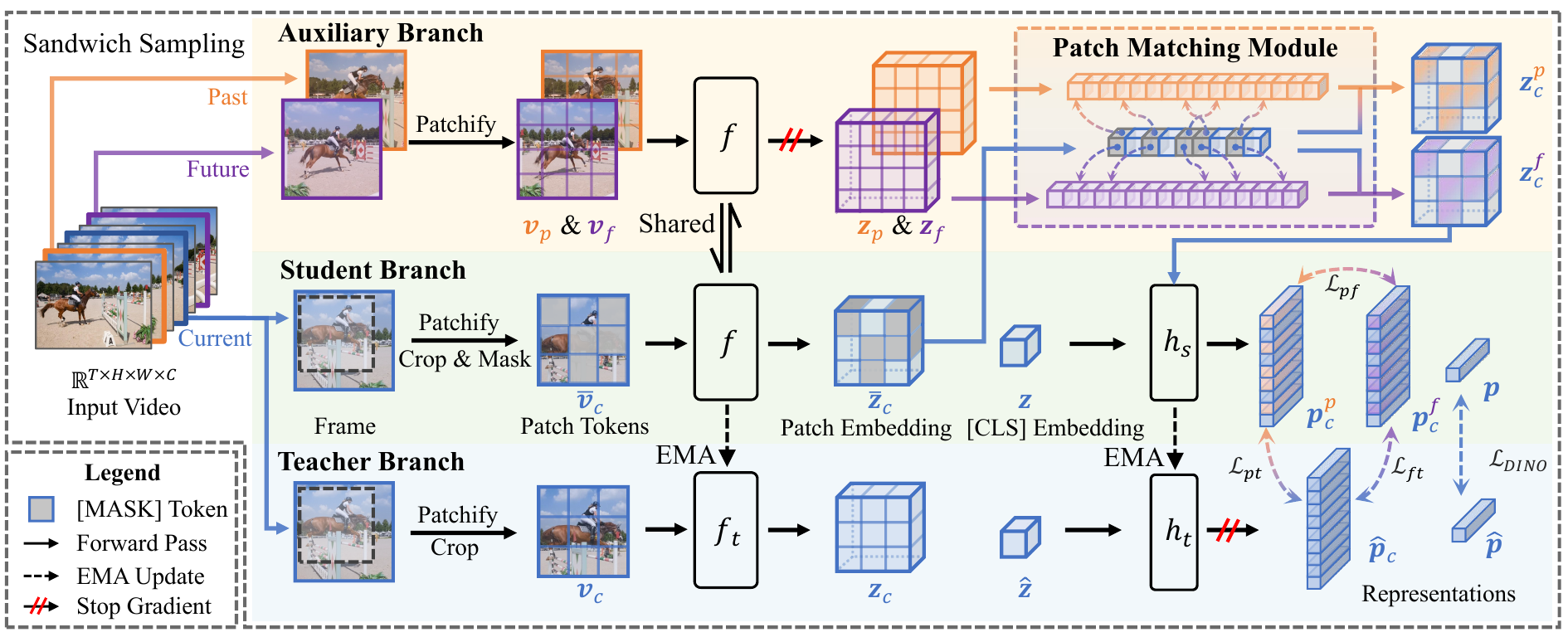}
    \caption{The architecture of our proposed framework. Using the sandwich sampling technique, we sample the past, current, and future frames from the given video. The current frame is fed into both the student and teacher branches, with a random mask applied to the student branch. The past and future frames are processed by an auxiliary branch to provide temporal correspondence through the Patch Matching Module. The reconstructed representations are then aligned with the teacher branch and pulled closer to reduce uncertainty.}
    \label{fig:pipeline}
    \vspace{-10pt}
\end{figure*}

\subsection{Self-supervised Video Representation Learning}
\label{subsec:related_work_SSL}
The rapid development of self-supervised learning has given rise to numerous video representation learning methods, broadly categorized into the following two paradigms: contrastive learning and masked modeling.

% \subsubsection{Contrastive Learning Method}
\textbf{Contrastive learning} leverages the inherent similarities and differences between samples to learn consistent representations, capturing global semantic information for effective distinguishing.
Early methods in image domain~\cite{SimCLRv1,SimCLRv2,MoCov1,MoCov2,MoCov3} utilize positive-negative pairs to learn discriminative features~\cite{ins_dis_1,ins_dis_2}, while later approaches eliminate negative samples to reduce complexity~\cite{BYOL,SwAV,SimSiam,Barlow_Twins}.
These methods have demonstrated favorable generalization capabilities~\cite{huangtowards}, leading to successful adaptation in the video domain~\cite{SlowFast}. Relying on spatiotemporal coherence~\cite{TCN,TiCo,CVRL}, semantic consistency~\cite{HiCo_1,HiCo_2,hua2024reconboost}, and motion clues~\cite{DSM,FAME} of the videos, different views can be generated for contrastive learning, allowing the model to excel in discriminative tasks like action recognition~\cite{UCF101,HMDB51,SlowFast} and video retrieval~\cite{FIVR,liu2024not}.

% \subsubsection{Masked Modeling Method}
\textbf{Masked modeling} captures co-occurrence relationships within local structures by restoring masked contents in the pixel space or latent space~\cite{SSL_survey,MIM_survey,MIM_Dark}.
\textbf{1) Pixel-space-oriented methods} typically involve regressing the raw RGB pixels of the masked patches through an encoder-decoder structure~\cite{MAE,SimMIM,I_JEPA,StoP,SiameseIM}.
By incorporating the temporal dimension, these approaches can be extended to masked video modeling methods~\cite{MAE_ST,VideoMAE,VideoMAEv2,BEVT,DropMAE,VideoMAC}, such as VideoMAE v2~\cite{VideoMAEv2}, which adopts extremely high masking ratios to reconstruct the masked video tubes, achieving impressive results on video-level tasks.
Recently, various sparse sampling methods have been proposed to reduce the computational burden of high-frame-rate video clips. SiamMAE~\cite{SiamMAE} samples two frames from a video, assigns asymmetric masks, and uses a Siamese network for restoration via a conditional decoder, promoting further improvements in the subsequent works. For instance, CropMAE~\cite{CropMAE} samples different augmented versions of a frame as input, and RSP~\cite{RSP} learns to predict the masked frame with prior and posterior distributions from a visible frame.
\textbf{2) Latent-space-oriented methods} aim to restore the masked regions into dense representations to reduce excessive focus on low-level visual details, enabling the model to maintain robust performance even in long-tailed scenarios~\cite{wen2024algorithm,dai2023drauc}.
Initially introduced in the image domain, methods such as iBOT~\cite{iBOT} and DINO v2~\cite{DINOv2} restore and align the representations in the latent space within a self-distillation framework, achieving notable performance across various tasks.
Motivated by the information compression principle~\cite{compression_1,compression_2,compression_3} from the above method, we propose a masked video modeling framework to restore the masked regions in the latent space, preserving high-level semantic information for downstream tasks.

\subsection{Temporal Correspondence Learning}
\label{subsec:related_work_Correspondence}

The stability and continuity of the visual world make temporal correspondence a natural supervisory signal~\cite{stability_1,stability_2,continuity,smooth}, enhancing visual representation for video-based downstream tasks such as action recognition~\cite{UCF101,HMDB51,SlowFast}, video segmentation~\cite{DAVIS17,VIP,wen2020dmvos,han2024aucseg}, and motion estimation~\cite{JHMDB}. 
Early methods for learning temporal correspondence rely on low-level features such as optical flow, frame residuals, motion vectors, and histograms of gradient~\cite{low_level_1,low_level_2,low_level_3,low_level_4}. To further capture spatiotemporal coherence, later methods took advantage of cycle consistency to learn robust visual representations~\cite{cycle_1,cycle_2,cycle_3,cycle_4}.
Additionally, more recent advances focused on refining deep features for object matching via contrastive or predictive learning~\cite{track_1,track_2,SiamMAE,track_3}.
In contrast, we incorporate temporal correspondence by introducing a sandwich strategy, which samples a past and a future frame to assist in restoring masked patches in the current frame, thus improving representations with more precise temporal alignment.

\section{Methodology}
\label{sec:method}

\subsection{Task Definition}
\label{subsec:task_defination}

An input video is represented as a sequence of frames $\bm{V} = \{\bm{v}_j \in \mathbb{R}^{H \times W \times C}\}_{j=1}^{T}$, where $H$, $W$, and $C$ denote the height, width, and channel of each frame respectively, and $T$ indicates the total number of frames in the video. For the sake of presentation, we denote the $t$-th time step frame as $\bm{V}(t)$. A frame $\bm{v}$ can be divided into $N = \lceil H\times W / p^2\rceil$ patches, each containing $p^2$ pixels. The goal of video representation learning is to obtain a representation for each patch in the video. Taking into account the generalization capability of pre-trained models, this work focuses on 
the frame-level encoder $f: \mathbb{R}^{H \times W \times C} \rightarrow \mathbb{R}^{N \times d}$, where $d$ is the embedding dimension. Compared with image representation learning, the frame-level encoder is trained on videos instead of isolated images, such that the encoder learns the dynamic features of objects from temporal information in a self-supervised manner. After pre-training, the encoder $f$ can be used to initialize various video downstream tasks.

In the mask modeling setting, we randomly sample some input patches and mask them with a learnable \texttt{[MASK]} token. Given a frame $\bm{v}$, denote the masked frame and the set of masked patches as $\bar{\bm{v}}$ and $\mathcal{M}(\bm{v})$, respectively. Then, we feed the masked video or frames into the encoder to restore the missing patches in the latent space or pixel space.

\subsection{Overview}

As illustrated in \cref{fig:pipeline}, our proposed framework contains three encoders with the same architecture: \textit{auxiliary} branch, \textit{student} branch, and \textit{teacher} branch. Given a training video, we sample three frames, which are designated as the past, current, and future frames. For the sake of presentation, the past and future frames are called \textit{auxiliary frames}.

This work aims to design a self-supervised video representation learning framework with two properties: \textbf{1)} the temporal information from auxiliary frames are aggregated to restore the masked patches in the current frame; \textbf{2)} the restoration process is conducted in the latent space instead of the pixel space.
To this end, as described in ~\cref{subsec:temporal_cor}, the masked current frame and the auxiliary frames are processed by the student branch and the auxiliary branch, respectively. Afterward, in ~\cref{subsubsec:patch_matching} we propose a \textit{Patch Matching Module} to retrieve the masked information from the auxiliary embeddings, yielding reconstructed embeddings of the current frame. Finally, as outlined in ~\cref{subsec:dual_frame}, we feed the current frame without mask into the teacher branch to generate supervision in a self-distillation manner.

\subsection{Uncertainty Reduction via Sandwich Sampling}
\label{subsec:temporal_cor}

As discussed in Sec.\ref{sec:introduction}, recent state-of-the-art video reconstruction methods primarily rely on a random sampling strategy to make the representations temporally predictive. This approach can be seen as training the model to reconstruct the current frame given the past frame, which can be challenging due to the infinite possibilities for the future.

Although predicting the future from the past is difficult, forecasting an intermediate point is much easier once the starting point and destination are known. Based on this idea, we propose a \textit{sandwich sampling strategy}. To reconstruct the masked current frame, we sample a past frame and a future frame to provide temporal guidance. Specifically, we first select the current frame $\vc = \bm{V}(t_c)$ at time step $t_c$ and obtain its masked version $\bar {\bm {v}}_c$. Then, we uniformly sample the time steps for the past and future frames within a predefined offset $[\alpha, \beta]$.

\begin{equation}
    \begin{aligned}
        t_p &\sim U(t_c - \beta \cdot T, t_c - \alpha \cdot T), \\
        t_f &\sim U(t_c + \alpha \cdot T, t_c + \beta \cdot T).
    \end{aligned}
\end{equation}

Next, the past and future frames are obtained as $\vp = \bm{V}(t_p)$ and $\vf = \bm{V}(t_f)$, respectively. These frames are passed through the encoder $f$ to obtain their representations: $\bar{\bm{z}}_c = f(\bar{\bm{v}}_c)$, $\zp = f(\vp)$, and $\zf = f(\vf)$.

\begin{figure}
  \centering
    \includegraphics[width=0.98\linewidth]{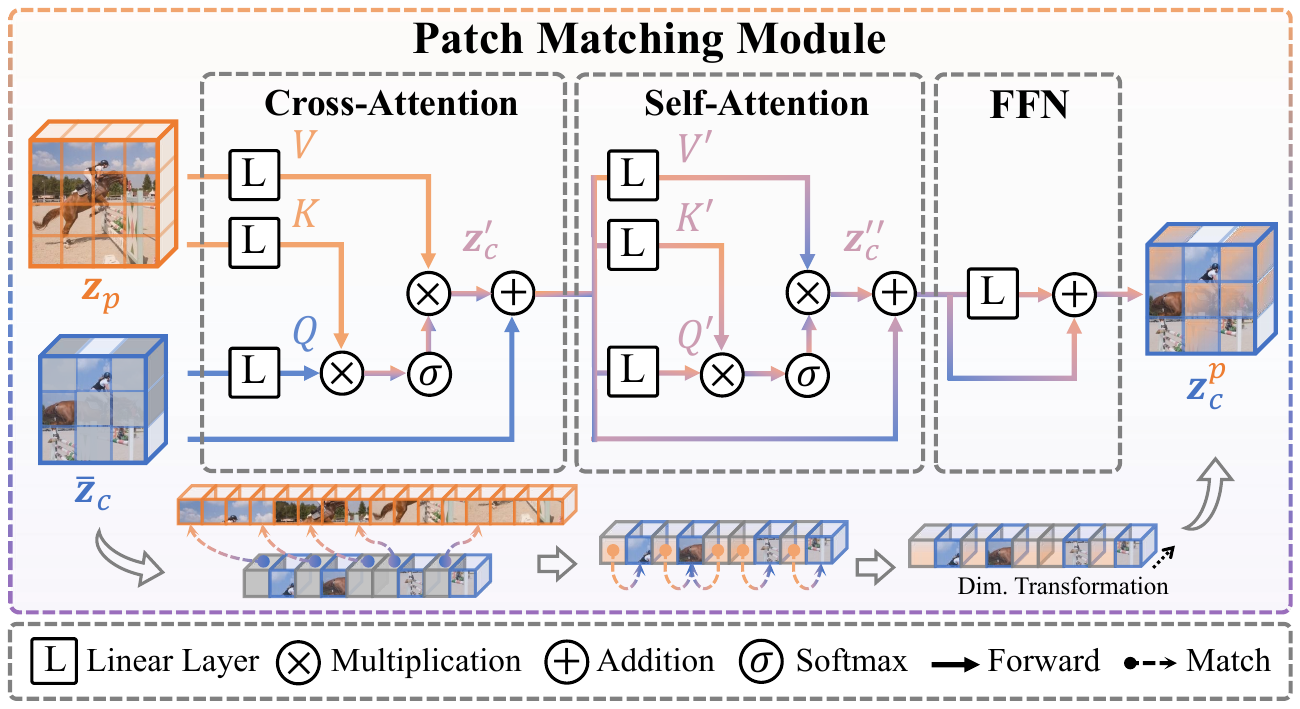}
    \caption{Structure of the Patch Matching Module. With the auxiliary of the past frame, $\zcsp = g(\zp, \bar{\bm z}_c)$ is generated via three blocks in PMM. The process is the same for the future frame.}
    \label{fig:PMM}
    \vspace{-10pt}
\end{figure}

\subsection{Patch Matching for Latent Correspondence}
\label{subsubsec:patch_matching}

In this subsection, we focus on establishing correspondence between the extracted frames. While previous works have attempted to build correspondence in the pixel space \cite{SiamMAE,RSP}, we argue that, due to the high redundancy of video information, these approaches fail to effectively compress task-irrelevant details. To address this issue, we propose finding the correspondence in a compact latent space. However, establishing precise correspondence between representations presents additional challenges due to the low resolution of the latent space. To overcome this, we introduce a Patch Matching Module (PMM) $g$ that treats the past and future frames as a dictionary and aims to retrieve the masked patches from it. As shown in \cref{fig:PMM}, the PMM consists of two main components: a cross-attention block for temporal information aggregation and a self-attention block to improve spatial correspondence.

Specifically, the first step is to retrieve masked patches from the adjacent frames. Using the masked patch representations $\bar{\bm{z}}_c$ as the query and the auxiliary representations $\zp$ as both the key and value, a cross-attention block is used to establish patch correspondence across the different frames:

\begin{equation}
    \zcs' = \text{softmax}_\tau \big(\frac{Q_c K_p^\top}{\sqrt{d}}\big) V_p ,
    \label{eq:cross-attn}
\end{equation}
where $Q_c=\bar{\bm{z}}_c W_q $, $K_p=\zp W_k $, $V_p=\zp W_v $ with $W_q, W_k, W_v \in \mathbb{R}^{d \times d}$ being weight matrices. The softmax function is applied along the feature dimension:

\begin{align}
    \text{softmax}_\tau(\bm x)^i = \frac{\exp(\bm x^i/\tau)}{\sum_{j=1}^d\exp(\bm x^j/\tau)},
\end{align}
$\tau > 0$ is a temperature hyperparameter. In \cref{eq:cross-attn}, the term $Q_c K_p^\top$ represents the similarity matrix. This approach allows the masked patches in the latent space to be filled with similar patches from the auxiliary frame. In contrast to previous methods that \textbf{imagine} the missing patches, our method \textbf{retrieves} them from temporal candidates.

Building upon this, to further enhance the local co-occurrence relationships between patches, $\zcs'$ is passed through a self-attention block for additional refinement:
\begin{equation}
    \zcs'' = \text{softmax}_\tau \big(\frac{Q_c' K_c'^\top}{\sqrt{d}}\big) V_c' ,
    \label{eq:self-attn}
\end{equation}
where $Q_c'=\zcs' W'_q $, $K_c'=\zcs' W'_k $, $V_c'=\zcs' W'_v $ with weight matrices $W'_q, W'_k, W'_v \in \mathbb{R}^{d \times d}$. Finally, a Feed Forward Network (FFN) is applied to adjust the representations in the latent space, enhancing their representational capacity:

\begin{equation}
    \zcsp = \text{FFN}(\zcs'') ,
    \label{eq:linear}
\end{equation}
and $\zcsf$ can be generated in the same manner.
Additionally, layer normalization and skip connections are applied to each block to ensure training stability. By combining these techniques, we construct our Patch Matching Module (PMM). In principle, multiple PMM modules can be stacked sequentially without requiring modifications to the network structure. In this work, we use one single module and apply a fully connected layer as the FFN.

\begin{algorithm}[tb]
\caption{T-CoRe Optimization Algorithm}
\label{alg:algo}
\LinesNumbered
\KwIn{Unlabeled dataset $\mathcal{D}$, maximum iterations $L$.}
\KwOut{Model parameters of student branch.}
Initialize model parameters.

\For{$l=1$ {\bfseries to} $L$}{
    Sample a batch of videos $\{\bm{V}_i\}_{i=1}^B$.

    \For{$k=1$ {\bfseries to} $B$}{
        \textcolor{cvprblue}{$\triangleright$ Sandwich Sampling}
        
        Select three frames $\vc$, $\vp$, and $\vf$ from $\bm{V}_k$.

        Get the masked current frame $\bar{\bm v}_c$.
    
        Extract dense representations $\bar{\bm z}_c$, $\zp$, $\zf$.

        \textcolor{cvprblue}{$\triangleright$ Patch Matching}

        Incorporate temporal correspondence via PMM: $\zcsp=g(\zp, \bar{\bm z}_c)$, $\zcsf=g(\zf, \bar{\bm z}_c)$.

        \textcolor{cvprblue}{$\triangleright$ Feature Reconstruction}
        
        Generating $\bm p_c^p$, $\bm p_c^f$ and $\hat {\bm p}_c$ by \cref{eq:proj}

        Distill representations from teacher branch by $\lpt$ and $\lft$ in \cref{eq:loss1}.

        Align $\bm p_c^p$ with $\bm p_c^f$ by $\lpf$ in \cref{eq:loss2}.

    }

    Update the student model with $\mathcal{L}$ (\cref{eq:total_loss}).

    Update the teacher model with EMA (\cref{eq:ema}).
    
}
\Return

\end{algorithm}
% \vspace{-10pt}

\subsection{Self-distillation-based Feature Reconstruction}
\label{subsec:dual_frame}

In the previous subsections, we have aggregated the temporal information from adjacent frames toward the representations of the current frame. In this subsection, we describe how we apply the loss function to these representations.

\noindent\textbf{Representation reconstruction.} 
We present our representation reconstruction loss, which is based on a self-distillation framework \cite{iBOT}. Specifically, we introduce a teacher model (denoted as $f_t$) to guide the student branch. Inspired by the widely used Mean Teacher technique \cite{tarvainen2017mean, MoCov1}, we update the teacher model by using the Exponential Moving Average (EMA) of the student weights:
\begin{align}
\label{eq:ema}
    \bm \theta_t \xleftarrow{} m \cdot \bm \theta_t + (1-m) \cdot \bm \theta ,
\end{align}
where $\bm \theta$ and $\bm \theta_t$ are the parameters of the student model $f$ and the teacher model $f_t$. As a temporal ensemble of the student model, the mean teacher is commonly used as a supervision signal in self-supervised learning \cite{BYOL, DINO, iBOT, MoCov3}.

To prevent model collapse \cite{jingunderstanding}, we also introduce projection heads $h$ and $h_t$ on top of the student and teacher models. Additionally, we apply centering and sharpening techniques \cite{DINO, iBOT} to the outputs of the projection heads:

\begin{equation}
    \begin{aligned}
        \bm p_c^p = \text{softmax}_\tau\left(h(\bm z_c^p)\right)&, ~
        \bm p_c^f = \text{softmax}_\tau\left(h(\bm z_c^f)\right), \\
        \hat {\bm p}_c = \text{softmax}_{\tau_t}&\left(h_t(f_t(\bm v_c)) - \bm \mu_t\right), 
        \label{eq:proj}
    \end{aligned}
\end{equation}
where $\tau_t > 0$ is the temperature of the softmax function and $\bm \mu_t$ is the empirical estimator of the mean of representations. Finally, our reconstruction loss is calculated as:

\begin{equation}
    \begin{aligned}
        \lpt &= -\frac{1}{|\mathcal M(\bm v_c)|}\sum_{i \in \mathcal{M}(\bm v_c)} (\hat {\bm p}_c)^i \log(\bm p_c^p)^i  , \\
        \lft &= -\frac{1}{|\mathcal M(\bm v_c)|}\sum_{i \in \mathcal{M}(\bm v_c)} (\hat {\bm p}_c)^i \log (\bm p_c^f)^i .
    \label{eq:loss1}
    \end{aligned}
\end{equation}

\noindent\textbf{Temporal squeezing.} 
Building on the concept of two-side squeezing, we introduce an additional constraint term on $\bm p_c^p$ and $\bm p_c^f$ to further reduce the uncertainty during training:
\begin{equation}
    \lpf = \frac{1}{|\mathcal M(\bm v_c)|}\sum_{i \in \mathcal{M}(\bm v_c)} \| (\bm p_c^p)^i - (\bm p_c^f)^i \|_2^2 .
\label{eq:loss2}
\end{equation}
Intuitively, without $\lpf$, the aggregated representations might overly depend on the past and future frames. As a result, both $\lpt$ and $\lft$ would be small, but $\bm p_c^p$ and $\bm p_c^f$ could exhibit a large divergence. This misalignment would introduce additional uncertainty and confuse the model.

\noindent\textbf{Frame-level alignment.}
Following prior works \cite{iBOT, DINOv2}, we combine frame-level losses with the patch-level losses described above to enable collaborative learning between global and local representations. Let ${\bm{p}}$ and ${\bm{z}}$ denote the sharpened and L2-normalized \texttt{[CLS]} embeddings of the student model, respectively. Denote $\hat{\bm{p}}$ as the centered and sharpened \texttt{[CLS]} embedding of the teacher model on $\vc$. We apply the DINO loss \cite{DINO} to align high-level semantics, and the KoLeo loss \cite{KoLeo} to promote diversity of representations within a batch:

\begin{align}
    \mathcal{L}_{DINO} &= -\frac 1 2 ({\bm{p}} \log \hat{\bm{p}}+\hat {\bm{p}} \log {\bm{p}}), \\
    \mathcal{L}_{koleo} &= - \frac 1 B \sum_{i=1}^B \log(\tilde d_{i}),
\end{align}
where $B$ denotes the batch size and $\tilde{d}_i = \min_{j \ne i} || {\bm{z}}_i - {\bm{z}}_j||_2$ is the distance between $ {\bm{z}}_i$ and its closest neighbor within the batch.

The total loss for video representation learning is then formulated in \cref{eq:total_loss}, where $\lambda_i$ are hyperparameters that control the contributions of each component.

\begin{equation}
    \mathcal{L} = \underbrace{\lambda_1(\lpt+\lft)+\lambda_2\lpf}_{patch-level} + \underbrace{\lambda_3\mathcal{L}_{DINO}+\lambda_4\mathcal{L}_{koleo}}_{frame-level}
\label{eq:total_loss}
\end{equation}

The overall optimization algorithm based on the methods presented in this section is summarized in \cref{alg:algo}. Note that the inner for-loop can be efficiently implemented using matrix operations to reduce computational cost.

\section{Experiments}
\label{sec:experiments}
\begin{table*}[htbp]
  \centering
  \vspace{-10pt}
  \resizebox{0.98\textwidth}{!}{%
    \begin{tabular}{cc|lcc|ccc|c|cc}
    \toprule
    \multicolumn{2}{c|}{\multirow{2}[2]{*}{Type}} & \multirow{2}[2]{*}{Method} & \multirow{2}[2]{*}{Backbone} & \multirow{2}[2]{*}{Epoch} & \multicolumn{3}{c|}{DAVIS-2017} & VIP   & \multicolumn{2}{c}{JHMDB} \\
          &   &       &       &       & $\mathcal{J}\&\mathcal{F}_{\mathrm{m}}$ & $\mathcal{J}_{\mathrm{m}}$   & $\mathcal{F}_{\mathrm{m}}$    & mIoU  & PCK@0.1 & PCK@0.2 \\
    \midrule
    \rowcolor[rgb]{ .867,  .922,  .969} \multicolumn{11}{c}{ImageNet-1k  Pre-trained} \\
    \midrule
    \multicolumn{2}{c|}{{Supervised}}
          & Supervised$^\dagger$~\cite{Supervised} $_\text{CVPR'16}$   & ResNet50 (26M) & 100   & 66.0  & 63.7  & 68.4  & 39.5  & 59.2  & 78.3  \\
    \cmidrule(lr){1-11}
    \multicolumn{2}{c|}{\multirow{3}[2]{*}{\centering \shortstack{Contrastive \\ Learning}}} 
          & DINO$^\dagger$~\cite{DINO} $_\text{ICCV'21}$  & ViT-S/16 (22M) & 800   & 61.8  & 60.2  & 63.4  & 36.2  & 45.6  & 75.0  \\
          & & ODIN$^\text{2}$$^\dagger$~\cite{ODIN2} $_\text{ECCV'22}$  & ResNet50 (26M) & 1000  & 54.1  & 54.3  & 53.9  & /     & /     & / \\
          & & CrOC$^\dagger$~\cite{Croc} $_\text{CVPR'23}$  & ViT-S/16 (22M) & 300   & 44.7  & 43.5  & 45.9  & 26.1  & /     & / \\
    \cmidrule(lr){1-11}
    \multicolumn{1}{c|}{\multirow{6}[1]{*}{\centering \shortstack{Masked \\ Modeling}}} & \multirow{3}[1]{*}{\centering \shortstack{Pixel \\ Space}} 
          & MAE$^\dagger$~\cite{MAE} $_\text{CVPR'22}$  & ViT-B/16 (87M) & 1600  & 53.5  & 52.1  & 55.0  & 28.1  & 44.6  & 73.4  \\
         \multicolumn{1}{c|}{} & & RC-MAE$^\dagger$~\cite{RC-MAE} $_\text{ICLR'23}$  & ViT-S/16 (22M) & 1600  & 49.2  & 48.9  & 50.5  & 29.7  & 43.2  & 72.3  \\
         \multicolumn{1}{c|}{} & & CropMAE$^\dagger$~\cite{CropMAE} $_\text{ECCV'24}$  & ViT-S/16 (22M) & 400   & 60.4  & 57.6  & 63.3  & 33.3  & 43.6  & 72.0  \\
    \cmidrule(lr){2-11}
    \multicolumn{1}{c|}{} & \multirow{3}[1]{*}{\centering \shortstack{Latent \\ Space}} 
          & iBOT$^\ddagger$~\cite{iBOT} $_\text{ICLR'22}$  & ViT-S/16 (22M) & 800   & 62.8  & 60.4  & \textbf{\textcolor{soft_blue}{65.3}}  & \textbf{\textcolor{soft_blue}{38.4}}  & 44.5  & 74.5  \\
         \multicolumn{1}{c|}{} & & DINO v2~\cite{DINOv2} $_\text{TMLR'24}$  & ViT-S/16 (22M) & 100   & \textbf{\textcolor{soft_blue}{63.2}}  & \textbf{\textcolor{soft_blue}{61.4}}  & 65.1  & 37.3  & \textbf{\textcolor{soft_red}{46.3}}  & \textbf{\textcolor{soft_blue}{75.4}}  \\
         \multicolumn{1}{c|}{} & & \textbf{T-CoRe (Ours)}      & ViT-S/16 (22M) & 100   &  \textbf{\textcolor{soft_red}{64.1}}  &  \textbf{\textcolor{soft_red}{62.1}}  &  \textbf{\textcolor{soft_red}{66.1}}  &  \textbf{\textcolor{soft_red}{39.7}}  &  \textbf{\textcolor{soft_blue}{46.2}}  & \textbf{\textcolor{soft_red}{75.5}} \\
    \midrule
    \rowcolor[rgb]{ .867,  .922,  .969} \multicolumn{11}{c}{Kinetics-400  Pre-trained} \\
    \midrule
     \multicolumn{2}{c|}{\multirow{3}[2]{*}{\centering \shortstack{Contrastive \\ Learning}}} 
          & SimCLR$^\dagger$~\cite{SimCLRv1} $_\text{ICML'20}$  & ViT-S/16 (22M) &  400  & 53.9  & 51.7  & 56.2  & 31.9  & 37.9  & 66.1  \\
          & & MoCo v3$^\dagger$~\cite{MoCov3} $_\text{ICCV'21}$  & ViT-S/16 (22M) &  400  & 57.7  & 54.6  & 60.8  & 32.4  & 38.4  & 67.6  \\
          & & DINO$^\dagger$~\cite{DINO} $_\text{ICCV'21}$  & ViT-S/16 (22M) &  400  & 59.5  & 56.5  & 62.5  & 33.4  & 41.1  & 70.3  \\
    \cmidrule(lr){1-11}
    \multicolumn{1}{c|}{\multirow{8}[1]{*}{\centering \shortstack{Masked \\ Modeling}}} & \multirow{5}[1]{*}{\centering \shortstack{Pixel \\ Space}} 
          & MAE$^\dagger$~\cite{MAE} $_\text{CVPR'22}$  & ViT-S/16 (22M) & 400 & 53.5  & 50.4  & 56.7  & 32.5  & 43.0  & 71.3  \\
         \multicolumn{1}{c|}{} & & VideoMAE$^\dagger$~\cite{VideoMAE} $_\text{NeurIPS'22}$  & ViT-S/16 (22M) & 1600  & 39.3  & 39.7  & 38.9  & 22.3  & 41.0  & 67.9  \\
         \multicolumn{1}{c|}{} & & SiamMAE$^\dagger$~\cite{SiamMAE} $_\text{NeurIPS'23}$  & ViT-S/16 (22M) & 400   & 57.9  & 56.0  & 60.0  & 33.2  & 46.1  & 74.0  \\
         \multicolumn{1}{c|}{} & & CropMAE$^\dagger$~\cite{CropMAE} $_\text{ECCV'24}$  & ViT-S/16 (22M) & 400   & 58.6  & 55.8  & 61.4  & 33.7  & 42.9  & 71.1  \\
         \multicolumn{1}{c|}{} & & RSP$^\dagger$~\cite{RSP}  $_\text{ICML'24}$  & ViT-S/16 (22M) & 400   & 60.1  & 57.4  & 62.8  & 33.8  & 44.6  & 73.4  \\
    \cmidrule(lr){2-11}
   \multicolumn{1}{c|}{} & \multirow{3}[1]{*}{\centering \shortstack{Latent \\ Space}} 
          & iBOT~\cite{iBOT} $_\text{ICLR'22}$  & ViT-S/16 (22M) & 400  &  \textbf{\textcolor{soft_blue}{63.5}} & 61.7  & \textbf{\textcolor{soft_blue}{65.4}}  & \textbf{\textcolor{soft_blue}{37.4}}  & 44.5  & 74.2  \\
         \multicolumn{1}{c|}{} & & DINO v2~\cite{DINOv2} $_\text{TMLR'24}$  & ViT-S/16 (22M) & 400   & 63.0  & \textbf{\textcolor{soft_blue}{62.0}}  & 64.1  & 37.0  & \textbf{\textcolor{soft_blue}{46.6}}  & \textbf{\textcolor{soft_blue}{74.8}}  \\ 
          % & \textbf{T-CoRe (Ours)}     & Latent & ViT-S/16 (22M) & 100   & 64.0  & 62.3  & 65.7  & 37.7  & 45.9  & 74.4  \\
         \multicolumn{1}{c|}{} & & \textbf{T-CoRe (Ours)}    & ViT-S/16 (22M) & 400   &  \textbf{\textcolor{soft_red}{64.7}}  &  \textbf{\textcolor{soft_red}{63.5}}  &  \textbf{\textcolor{soft_red}{66.0}}  &  \textbf{\textcolor{soft_red}{37.8}}  &  \textbf{\textcolor{soft_red}{47.0}}  & \textbf{\textcolor{soft_red}{75.2}} \\

    \bottomrule
    \end{tabular}%
    }
    \vspace{-2pt}
    \caption{Comparison with prior methods on three dense-level video downstream tasks. Results marked with $^\dagger$ are sourced directly from previous studies, while $^\ddagger$ denotes evaluations conducted based on the officially provided pre-trained weights. Missing values represent the absence of reported results or implementations. The best and second-best results are highlighted in \textbf{\textcolor{soft_red}{soft red}} and \textbf{\textcolor{soft_blue}{soft blue}}, respectively.}
  \label{tab:main_results}%
  \vspace{-6pt}
\end{table*}%

\begin{table}[htbp]
  \centering
  \setlength{\tabcolsep}{3pt}
  \resizebox{0.48\textwidth}{!}{%
    \begin{tabular}{cc|ccc|c|cc}
    \toprule
    \multirow{2}[2]{*}{Method}  & \multirow{2}[2]{*}{Epoch} & \multicolumn{3}{c|}{DAVIS-2017} & VIP   & \multicolumn{2}{c}{JHMDB} \\
              &            & $\mathcal{J}\&\mathcal{F}_{\mathrm{m}}$ & $\mathcal{J}_{\mathrm{m}}$   & $\mathcal{F}_{\mathrm{m}}$    & mIoU  & PCK@0.1 & PCK@0.2 \\
          \midrule
          \rowcolor[rgb]{ .867,  .922,  .969} \multicolumn{8}{c}{Kinetics-400  Pre-trained} \\
          \midrule
           MAE-ST$^\dagger$~\cite{MAE_ST}    & 1600  & 54.6  & 55.5  & 53.6  & 33.2  & 44.4  & 72.5  \\
           DropMAE$^\dagger$~\cite{DropMAE}  & 1600  & 53.4  & 51.8  & 55.0  & 31.1  & 42.3  & 69.2  \\
           % SiamMAE$^\dagger$~\cite{SiamMAE}    & 2000  & 62.8  & 60.9  & 64.6  & 38.4  & 47.2  & 76.4  \\
           % SiamMAE~\cite{SiamMAE}    & 400  &   &   &   &   &  &   \\
           CropMAE~\cite{CropMAE}    & 400  &  57.8 & 56.9  & 58.7  & 33.0  & 45.3 & 73.3  \\
           RSP$^\dagger$~\cite{RSP}  & 400   & 60.5  & 57.8  & 63.2  & 34.0  & 46.0  & 74.6  \\
           DINO v2~\cite{DINOv2}  & 200   & \textbf{\textcolor{soft_blue}{64.9}}  & \textbf{\textcolor{soft_blue}{63.3}}  & \textbf{\textcolor{soft_blue}{66.6}}  & \textbf{\textcolor{soft_blue}{38.2}}  & \textbf{\textcolor{soft_blue}{46.8}}  & \textbf{\textcolor{soft_blue}{75.8}}  \\ 
           \textbf{T-CoRe (Ours)}  & 200  & \textbf{\textcolor{soft_red}{66.4}}  & \textbf{\textcolor{soft_red}{64.6}}  &  \textbf{\textcolor{soft_red}{68.2}}  & \textbf{\textcolor{soft_red}{38.9}}  &  \textbf{\textcolor{soft_red}{47.1}} & \textbf{\textcolor{soft_red}{75.8}} \\
    \bottomrule
    \end{tabular}%
    }
    \caption{Comparison with prior methods with larger backbones (ViT-L/16 for MAE-ST and ViT-B/16 for the others). Results marked with $^\dagger$ are sourced directly from previous studies.}
  \label{tab:main_results_large_backbone}%
  \vspace{-12pt}
\end{table}%

In this section, we conduct experiments to validate the effectiveness of our T-CoRe. We begin with the basic settings in \cref{subsec:experiment_setup}. Next, we provide quantitative comparisons on three downstream tasks in \cref{subsec:quantitative_results}. Finally, ablation studies and qualitative results are shown in \cref{subsec:ablation_study} and \cref{subsec:qualitative_results}.
Please refer to \textit{Supplementary Material} for further details.

\subsection{Experiment Setup}
\label{subsec:experiment_setup}

\noindent\textbf{Pre-training.} 
We adopt ViT-Small and ViT-Base~\cite{Vision_Transformer} with a patch size of 16 as the backbone models in our framework. 
For pre-training on Kinetics-400~\cite{Kinetics}, we first extract frames from each video with $FPS=2$. In each epoch, three subsequent frames are randomly selected according to the \textit{past/future offset} with $\alpha = 0.15$ and $\beta = 0.25$ at default. Then, 2 global views ($224\times 224$) and 8 local views ($96\times 96$) are generated from the \textit{current frame}. Following ~\cite{DINOv2}, $50\%$ of the global crops in the student branch are masked with $10\% \sim 50\%$ of randomly selected patches.
We also extend our method to pre-train on the ImageNet-1k~\cite{Imagenet}, where the $k$-NN images are applied to simulate the adjacent frames. In this setting, the auxiliary branch receives one $k$-NN image where $k$ is set to $5$ in our setting.

\noindent\textbf{Optimizing strategies.} 
During training, if not restated, the ViT-S/16 model is trained on Kinetics-400 for 400 epochs with a batch size of $256$ ($128$ for the ViT-B/16 model). The student branch is optimized by AdamW~\cite{AdamW}. The learning rate for the student branch \textit{lr} is set to $1\times 10^{-3}$ and decay to $1\times 10^{-6}$ with a cosine decay schedule. For the PMM, the learning rate is set to $0.1 \times \textit{lr}$. The hyper-parameters of the loss function are set as follows: $\lambda_1 = 0.8$, $\lambda_2 = 20$, $\lambda_3 = 1.0$, and $\lambda_4 = 0.1$. Note that $\lambda_3$ and $\lambda_4$ are taken from the default setting of \cite{DINOv2} and we \textbf{only tune the $\lambda_1$ and $\lambda_2$} in our experiments.

\noindent\textbf{Evaluation protocol.} 
We evaluate our framework using three dense-level downstream tasks: video object segmentation on DAVIS-2017~\cite{DAVIS17}, semantic part propagation on VIP~\cite{VIP}, and human pose propagation on JHMDB~\cite{JHMDB}. 
Following the previous works~\cite{SiamMAE,CropMAE,RSP}, all tasks are evaluated using a semi-supervised protocol, where the pre-trained model is provided with the ground-truth segmentation mask of the first frame and required to propagate the mask across the rest frames of the video without fine-tuning.

\noindent\textbf{Competitors.} We compare T-CoRe with various state-of-the-art self-supervised representation learning methods that categorized into two types: \textbf{1) }\textit{Contrastive learning methods}, including SimCLR~\cite{SimCLRv1}, MoCo v3~\cite{MoCov3}, DINO~\cite{DINO}, ODIN$^\text{2}$~\cite{ODIN2}, and CrOC~\cite{Croc}; \textbf{2) }\textit{Masked modeling methods}, including MAE~\cite{MAE}, MAE-ST~\cite{MAE_ST}, RC-MAE~\cite{RC-MAE}, VideoMAE~\cite{VideoMAE}, DropMAE~\cite{DropMAE}, SiamMAE~\cite{SiamMAE}, CropMAE~\cite{CropMAE}, RSP~\cite{RSP}, iBOT~\cite{iBOT} and DINO v2~\cite{DINOv2}. For a fair comparison, we do not apply the costly techniques in DINO v2~\cite{DINOv2} like distillation from a larger model or pre-training with a larger resolution.

\subsection{Main Results on Downstream Tasks}
\label{subsec:quantitative_results}

The evaluation results on three downstream tasks are reported in \cref{tab:main_results,tab:main_results_large_backbone}, leading to the following conclusions:
\textbf{1)} Our proposed T-CoRe consistently outperforms all the competitors across tasks, metrics, and model architectures. These results highlight the effectiveness of our temporal correspondence learning framework.
\textbf{2)} Comparing with existing pixel-level masked modeling methods, the proposed T-CoRe gets improvements of at least 4.6\%, 4.0\%, and 1.2\% on $\mathcal{J}\&\mathcal{F}_{\mathrm{m}}$, mIoU, and PCK@0.2, respectively. These outcomes emphasize the effectiveness of aligning representations in the \textbf{latent space}, which provides more compact representations for downstream tasks.
\textbf{3)} Compared to advanced self-distillation frameworks like iBOT \cite{iBOT} and DINO v2 \cite{DINOv2}, which are limited by spatial alignment, T-CoRe achieves an increment of $1.2\%\sim1.7\%$ on $\mathcal{J}\&\mathcal{F}_{\mathrm{m}}$. Benefiting from our proposed PMM and sandwich sampling technique, temporal information is aggregated via the auxiliary branch. This facilitates representations with more precise spatiotemporal structures, eventually leading to superior performances on the downstream tasks.
\textbf{4)} The extensive experiments on ImageNet-1k and larger backbone demonstrates the scalability of T-CoRe. It is noteworthy that with the same or fewer training epochs, T-CoRe reaches the state-of-the-art performance and even exceeds the \textbf{supervised method} on the VIP dataset. This suggests its potential as a more general self-supervised pre-training framework on wider applications of video understanding.

\subsection{Ablation Study}
\label{subsec:ablation_study}

\begin{table*}[t]
\vspace{-10pt}
  \centering
  \begin{minipage}{0.66\textwidth}
    \centering
    \subfloat[Ablation on components of T-CoRe. \label{tab:ablation_aux}]{
        \begin{minipage}{0.46\linewidth}
            \centering
            \setlength{\tabcolsep}{4.4pt}
            \resizebox{1.0\textwidth}{!}{
                \begin{tabular}{ccccc|ccc}
                \toprule
                \multicolumn{1}{c}{\multirow{2}[2]{*}{No.}} & \multirow{2}[2]{*}{\centering \shortstack{Use \\ Aux.}} & \multicolumn{1}{c}{\multirow{2}[2]{*}{$\lpt$}} & \multicolumn{1}{c}{\multirow{2}[2]{*}{$\lft$}} & \multicolumn{1}{c|}{\multirow{2}[2]{*}{$\lpf$}} & \multicolumn{3}{c}{DAVIS-2017}  \\
                &      &       &       &       & $\mathcal{J}\&\mathcal{F}_{\mathrm{m}}$ & $\mathcal{J}_{\mathrm{m}}$   & $\mathcal{F}_{\mathrm{m}}$ \\
                \midrule
                1 &  &      &      &      & 61.9  & 59.7  & 64.0   \\
                2 & \ding{51} & \ding{51} &  &  & 63.4  & 61.4  & \underline{65.3}  \\
                3 & \ding{51} &  & \ding{51} &  & 63.0  & 61.0  & 65.1 \\
                4 & \ding{51} & \ding{51} & \ding{51} &  & \underline{63.5}  & \underline{62.0} & 65.0  \\
                5 & \cellcolor[rgb]{ .886,  .937,  .855} \ding{51} & \cellcolor[rgb]{ .886,  .937,  .855} \ding{51} & \cellcolor[rgb]{ .886,  .937,  .855} \ding{51} & \cellcolor[rgb]{ .886,  .937,  .855} \ding{51} & \cellcolor[rgb]{ .886,  .937,  .855} \textbf{64.4} & \cellcolor[rgb]{ .886,  .937,  .855} \textbf{63.0}  & \cellcolor[rgb]{ .886,  .937,  .855} \textbf{65.9}   \\
                \bottomrule
                \end{tabular}%
            }
        \end{minipage}
}
    % \hspace{0.1cm}
    \subfloat[Ablation on positional alignment. \label{tab:ablation_pos}]{
        \begin{minipage}{0.48\linewidth}
            \centering
            \setlength{\tabcolsep}{5.5pt}
            \resizebox{1.0\textwidth}{!}{
                \begin{tabular}{cccc|ccc}
                \toprule
                \multicolumn{1}{c}{\multirow{2}[2]{*}{No.}} & \multirow{2}[2]{*}{\centering \shortstack{Use \\ Aux.}} & \multicolumn{1}{c}{\multirow{2}[2]{*}{\centering \shortstack{Space \\ Fixed}}} & \multicolumn{1}{c|}{\multirow{2}[2]{*}{\centering \shortstack{Time \\ Fixed}}} & \multicolumn{3}{c}{DAVIS-2017}  \\
                  &  &   &    & $\mathcal{J}\&\mathcal{F}_{\mathrm{m}}$ & $\mathcal{J}_{\mathrm{m}}$   & $\mathcal{F}_{\mathrm{m}}$  \\
                \midrule
                1 &   &   &      & 55.6  & 53.4  &  57.7  \\
                2 &   & \ding{51}   &  & 57.9  & 55.4  & 60.4    \\
                3 &   &  & \ding{51}   & 57.9  & 55.5  & 60.3    \\
                4 &   &  \ding{51} & \ding{51} & 61.9  & 59.7  & 64.0   \\
                5 & \cellcolor[rgb]{ .886,  .937,  .855} \ding{51}  & \cellcolor[rgb]{ .886,  .937,  .855} \ding{51} & \cellcolor[rgb]{ .886,  .937,  .855} \ding{51}  & \cellcolor[rgb]{ .886,  .937,  .855} \textbf{64.4} & \cellcolor[rgb]{ .886,  .937,  .855} \textbf{63.0}  & \cellcolor[rgb]{ .886,  .937,  .855} \textbf{65.9}   \\
                \bottomrule
                \end{tabular}%
            }
        \end{minipage}
}
    \par\vspace{0.2cm}
    \subfloat[Ablation on sampling strategy. \label{tab:ablation_sample}]{
        \begin{minipage}{0.48\linewidth}
            \centering
            \setlength{\tabcolsep}{4pt}
            \resizebox{1.0\textwidth}{!}{
                \begin{tabular}{ccc|ccc}
                \toprule
                \multicolumn{1}{c}{\multirow{2}[2]{*}{No.}} & \multicolumn{1}{c}{\multirow{2}[1]{*}{\centering \shortstack{Past \\ Offset}}} & \multicolumn{1}{c|}{\multirow{2}[1]{*}{\centering \shortstack{Future \\ Offset}}} & \multicolumn{3}{c}{DAVIS-2017}  \\
                &     & \multicolumn{1}{c|}{}  & $\mathcal{J}\&\mathcal{F}_{\mathrm{m}}$ & $\mathcal{J}_{\mathrm{m}}$   & $\mathcal{F}_{\mathrm{m}}$ \\
                \midrule
                1 & $\bm{-} [0.05, 0.15]$ & $\bm{+} [0.05, 0.15]$ & 64.0  & 62.3 & 65.7  \\
                2 & $\bm{-} [0.10, 0.20]$ & $\bm{+} [0.10, 0.20]$ & \underline{64.2} &\underline{62.4} & \textbf{66.0}  \\
                3 & \cellcolor[rgb]{ .886,  .937,  .855} $\bm{-} [0.15, 0.25]$ & \cellcolor[rgb]{ .886,  .937,  .855} $\bm{+} [0.15, 0.25]$ & \cellcolor[rgb]{ .886,  .937,  .855} \textbf{64.4} & \cellcolor[rgb]{ .886,  .937,  .855} \textbf{63.0}  & \cellcolor[rgb]{ .886,  .937,  .855} \underline{65.9}  \\
                4 & $\bm{-} [0.20, 0.30]$ & $\bm{+} [0.20, 0.30]$  & 62.8 & 61.1 & 64.5  \\
                5 & $\bm{-} [0.15, 0.25]$ & $\bm{-} [0.15, 0.25]$  & 63.4 & 61.7 & 65.1  \\
                6 & $\bm{+} [0.15, 0.25]$ & $\bm{+} [0.15, 0.25]$  & 63.3 & 61.3 & 65.3  \\
                \bottomrule
                \end{tabular}%
            }
        \end{minipage}
}
    % \hspace{0.1cm}
    \subfloat[Ablation on the masking scheme. \label{tab:ablation_mask}]{
        \begin{minipage}{0.48\linewidth}
            \centering
            \setlength{\tabcolsep}{3.3pt}
            \resizebox{1.0\textwidth}{!}{
                \begin{tabular}{cccc|ccc}
                \toprule
                \multicolumn{1}{c}{\multirow{2}[1]{*}{No.}} & \multicolumn{1}{c}{\multirow{2}[1]{*}{\centering \shortstack{Mask \\ Prob.}}} & \multicolumn{1}{c}{\multirow{2}[1]{*}{\centering \shortstack{Min. Mask \\  Ratio}}} & \multicolumn{1}{c|}{\multirow{2}[1]{*}{\centering \shortstack{Max. Mask \\ Ratio}}} & \multicolumn{3}{c}{DAVIS-2017}  \\
                &      &       &       & $\mathcal{J}\&\mathcal{F}_{\mathrm{m}}$ & $\mathcal{J}_{\mathrm{m}}$   & $\mathcal{F}_{\mathrm{m}}$  \\
                \midrule
                1     & 0.1   & 0.1 & 0.5 &  60.1  &  58.1  &  62.2  \\
                2     & 0.3 & 0.1 & 0.5  &  63.5 & 61.8  & 65.3  \\
                3     & \cellcolor[rgb]{ .886,  .937,  .855} 0.5   & \cellcolor[rgb]{ .886,  .937,  .855}0.1 & \cellcolor[rgb]{ .886,  .937,  .855}0.5 & \cellcolor[rgb]{ .886,  .937,  .855}\textbf{64.4} & \cellcolor[rgb]{ .886,  .937,  .855}\textbf{63.0}  & \cellcolor[rgb]{ .886,  .937,  .855}\textbf{65.9}   \\
                4     & 0.7   & 0.1 & 0.5 &  62.7  & 61.3  & 64.2 \\
                5     & 0.5   & 0.2 & 0.6 &  \underline{64.1}  &  \underline{62.3}  & \textbf{65.9} \\
                6     & 0.5   & 0.3 & 0.7 &  63.7  &  61.9  & \underline{65.5} \\
                \bottomrule
                \end{tabular}%
            }
        \end{minipage}
}
    % \vspace{-0.3cm}
    \caption{Ablation results on DAVIS-2017. Default settings are highlighted in \colorbox[rgb]{ .886,  .937,  .855}{green}. The best and second best results are marked with \textbf{bold} and \underline{underline}.} \label{tab:ablation}
    \end{minipage}
    \hspace{0.1cm}
    \begin{minipage}{0.32\textwidth}
    \centering
    \includegraphics[width=0.96\textwidth]{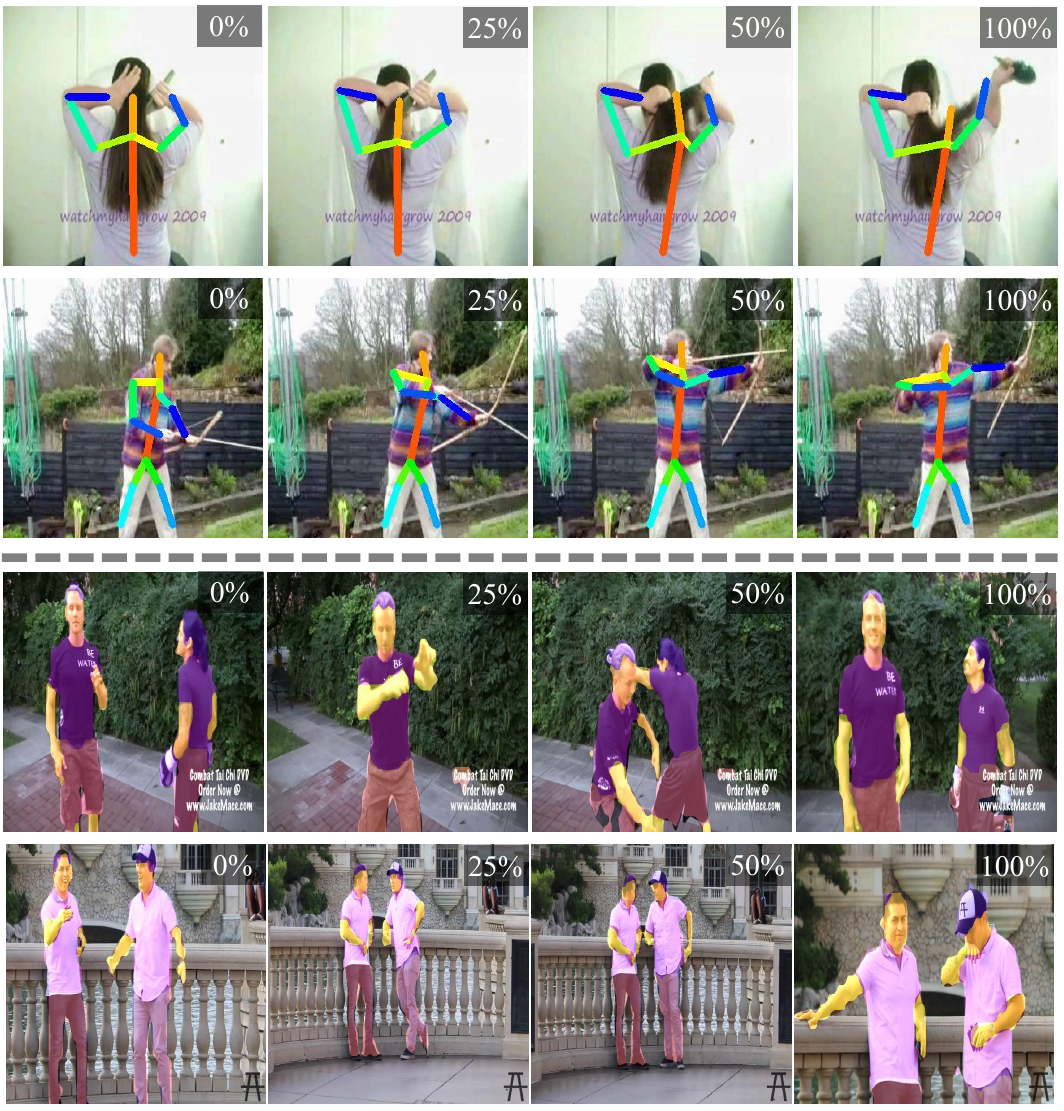}
    \captionof{figure}{Visualization results of T-CoRe for human pose propagation on JHMDB (\textit{top}) and body part propagation on VIP (\textit{bottom}).}
    \label{fig:vis_results}
    \end{minipage}
    \vspace{-1pt}
    % \vspace{-0.47cm}
\end{table*}
\vspace{-4pt}

\begin{figure*}
% \vspace{-10pt}
  \centering
    \includegraphics[width=0.86\linewidth]{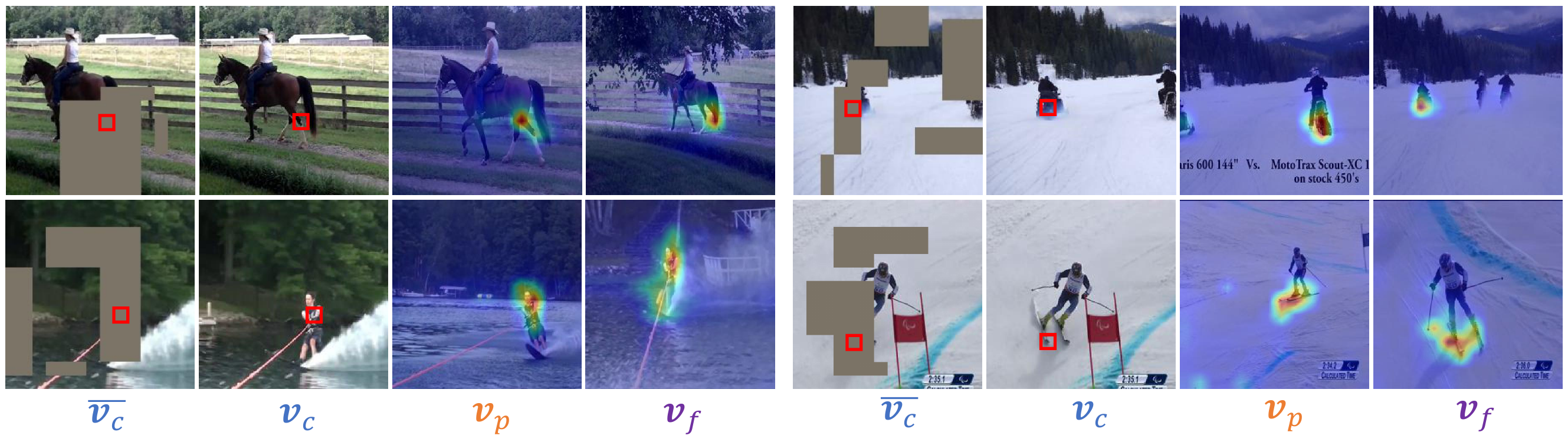}
    \vspace{-4pt}
    \caption{Cross-attention heatmaps of the masked current frame $\bar{\bm v}_c$ to the past and future frames $\vp, \vf$ in the PMM. Best viewed in colors.}
    \label{fig:vis_heatmaps}
    \vspace{-8pt}
\end{figure*}

In this part, we first investigate how the components of our framework affect the performance and then explore the impact of the sampling strategy and masking ratio.
To reduce the computation cost of ablation experiments, we adopt a swift protocol: Pre-training for 100 epochs with ViT-S/16 backbone on Kinetics-400 and evaluating on DAVIS-2017. The evaluation results are reported in \cref{tab:ablation}.

\noindent\textbf{Effect of auxiliary branch.}
The comparison between \textit{line 1} and \textit{lines 2-3} in \cref{tab:ablation_aux} reveals that introducing either a past or future frame in the auxiliary branch to guide the reconstruction of masked patches of the current frame improves $\mathcal{J}\&\mathcal{F}_{\mathrm{m}}$ by 1.5\% and 1.1\%, respectively. Such improvements emphasize the necessity of integrating temporal correspondence. The slightly lower results with the future frame may be attributed to the anisotropy property of the time~\cite{SiamMAE}. 
Furthermore, comparing \textit{lines 2-3} and \textit{line 4} shows that simply using both past and future frames for auxiliary only yields marginal improvement. However, as seen in \textit{line 5}, aligning the two restored representations guided by both past and future frames via $\lpf$ further boosts performance by 1.0\% on average, demonstrating the effectiveness of our proposed auxiliary branch with the PMM.

\noindent\textbf{Effect of strict positional alignment.}
A simple way for self-distillation models to learn spatiotemporal structure is to provide different crops for the student and teacher branches.
However, comparing \textit{lines 1-4} in \cref{tab:ablation_pos} shows that using different spatial or temporal crops reduces performance obviously. This is because the dense representations require strict positional alignment in such a restoration task. Unfortunately, fixing the spatial and temporal positions prevents the model from capturing temporal information. As shown in \textit{line 5}, introducing the auxiliary branch into the self-distillation architecture results in 2.6\% improvement on average, which highlights the effectiveness of the auxiliary branch for integrating temporal correspondence.

\noindent\textbf{Impact of sampling strategy.}
In \cref{tab:ablation_sample}, \textit{lines 1-4} present the effect of different temporal offsets between the current frame and the past/future frames. Performance decreases when the past/future frames are either too close or too distant to the current frame.
Moreover, as shown in \textit{line 5-6}, there is also a decline when both auxiliary frames are either past or future frames. This indicates that simply increasing the number of frames is insufficient, whereas incorporating comprehensive temporal context to reduce uncertainty enables more accurate representations. 
% These outcomes demonstrate the effectiveness of the proposed sandwich sampling strategy.

\noindent\textbf{Impact of the masking scheme.}
In \cref{tab:ablation_mask}, \textit{lines 1-4} examine the impact of various masking probabilities, while \textit{lines 3,5,6} evaluate different masking ratios. The performance declines with fewer masked patches since the task degrades into relying on the unmasked regions within the current frame itself for reconstruction. Similarly, excessive masking increases the complexity of patch matching and restoration, also resulting in performance decline.
These results underscore the importance of selecting the appropriate masking scheme to balance the reconstruction difficulty.

\subsection{Qualitative Results}
\label{subsec:qualitative_results}
We provide several qualitative results to intuitively validate the performance of our proposed framework. More results are available in \textit{Supplementary Material}.

\noindent\textbf{Visualization of downstream tasks.}
In \cref{fig:vis_results}, we present several prediction results on downstream tasks, where T-CoRe demonstrates favorable overall performance.

\noindent\textbf{Attention maps.}
In \cref{fig:vis_heatmaps}, we visualize the cross-attention maps of the masked patches between the current frame and both the past and future frames in PMM.
As shown in the figure, even if the queried patch is masked in the current frame, the proposed PMM could successfully match the correspondence patch in the auxiliary frames, indicating a strong ability to establish temporal correspondence.

\section{Conclusion}
\label{sec:conclusion}

In this work, we propose a self-supervised framework for video representation learning, which leverages temporal correspondence for representation restoration in latent space.
Within this framework, we design a sandwich sampling strategy to reduce reconstruction uncertainty and introduce an auxiliary branch into a self-distillation architecture to generate high-level semantic representations with temporal information.
Experimental results demonstrate that our framework consistently surpasses existing methods across several downstream tasks.

\section*{Acknowledgments}
This work was supported in part by the National Key R\&D Program of China under Grant 2018AAA0102000, in part by National Natural Science Foundation of China: 62236008, 62441232, U21B2038, U23B2051 and 62122075, in part by Youth Innovation Promotion Association CAS, in part by the Strategic Priority Research Program of the Chinese Academy of Sciences, Grant No. XDB0680201.

{
    \small
    \bibliographystyle{ieeenat_fullname}
    \bibliography{main}
}


\newpage
\appendix
\setcounter{page}{1}
\maketitlesupplementary

\section*{{\Large{Contents}}}
\startcontents[appendices]
\printcontents[appendices]{l}{1}{\setcounter{tocdepth}{3}}

\section{Supplementary Explanation of Method}
\subsection{Differences with Previous Methods}

\begin{figure*}
% \vspace{-10pt}
  \centering
  \begin{subfigure}{0.33\linewidth}
    \includegraphics[width=1.0\linewidth]{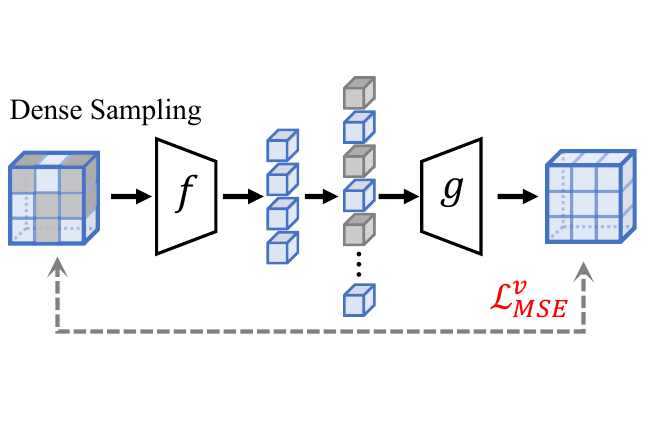}
    \caption{Structure of early MVM methods.}
    \label{fig:compare_methods_1}
    \end{subfigure}
  % \hfill
  \begin{subfigure}{0.33\linewidth}
    \includegraphics[width=1.0\linewidth]{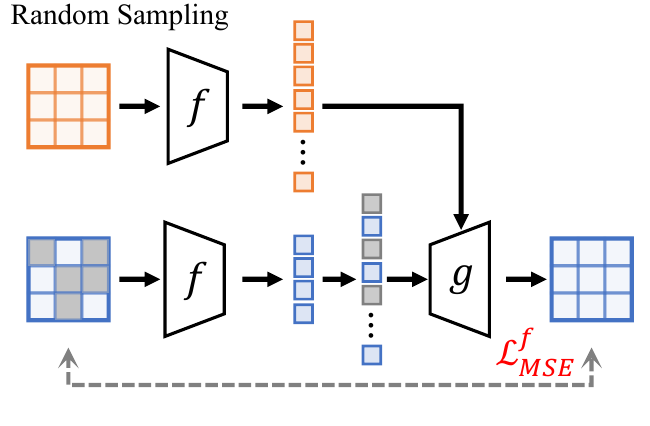}
    \caption{ Structure of recent MVM methods.}
    \label{fig:compare_methods_2}
  \end{subfigure}
  % \hfill
  \begin{subfigure}{0.33\linewidth}
    \includegraphics[width=1.0\linewidth]{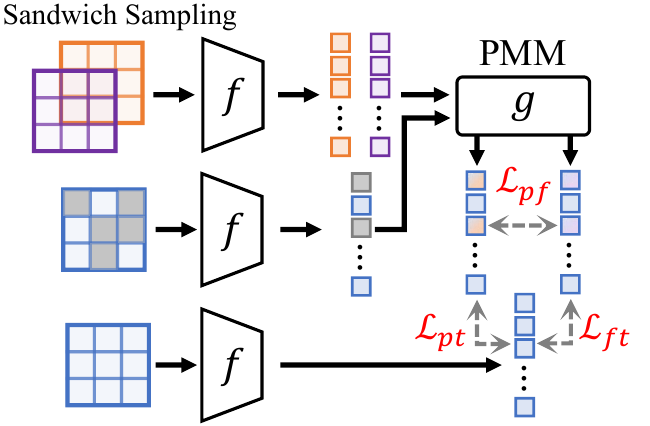}
    \caption{ Structure of T-CoRe (Ours).}
    \label{fig:compare_methods_3}
  \end{subfigure}
    \caption{Comparison of structures between our framework with previous MVM methods.}
    \label{fig:compare_methods}
    \vspace{-6pt}
\end{figure*}

The success of Masked Image Modeling (MIM) has inspired its extension to the video domain as Masked Video Modeling (MVM), which has demonstrated impressive performance for multiple video tasks. 

Given a video $\bm{V} = \{\bm{v}_j \in \mathbb{R}^{H \times W \times C}\}_{j=1}^{T}$, early MVM methods~\cite{MAE_ST,VideoMAE,VideoMAEv2,BEVT,DropMAE,MaskViT,OmniMAE} aim to train the video encoder $f$ through dense sampling from videos and forcing the model to recover the masked pixels. Specifically, as shown in \cref{fig:compare_methods_1}, for a sampled video clip $\bm{V}' \subset \bm{V}$, $\bm{V}'$ is divided into several tubes, which will be masked with learnable \texttt{[MASK]} tokens at a large probability. Next, the masked clip is passed through an encoder-decoder structure to predict the original clip $\widehat{\bm{V}'}$ by recovering the masked pixels. Finally, the mean square error (MSE) loss is utilized to minimize the difference between $\widehat{\bm{V}'}$ and $\bm{V}'$:
\begin{equation}
    \mathcal{L}_{MSE}^v = \frac{1}{|\mathcal{M}|}\sum_{i \in \mathcal{M}} \| (\bm{V}')^i - (\widehat{\bm{V}'})^i \|_2^2 ,
\label{eq:mse_loss_1}
\end{equation}
where $\mathcal{M}$ is the set of \texttt{[MASK]} tokens.

Recent MVM methods~\cite{SiamMAE,CropMAE,STP,RSP} have shifted to a random sampling strategy to reduce computational cost in the early efforts. Typically, as shown in \cref{fig:compare_methods_2}, these methods first sample a current frame $\bm{v}_c \in \bm{V}$ with several masked patches, then sample an unmasked past frame $\bm{v}_p \in \bm{V}$. The masked current frame $\bm{v}_c$ are restored to $\widehat{\bm{v}}_c$ using $\bm{v}_p$ as prior information based on a conditional decoder. Similarly, the mean square error (MSE) loss is used to minimize the difference between $\widehat{\bm{v}}_c$ and $\bm{v}_c$:
\begin{equation}
    \mathcal{L}_{MSE}^f = \frac{1}{|\mathcal{M}|}\sum_{i \in \mathcal{M}} \| (\bm{v}_c)^i - (\widehat{\bm{v}}_c)^i \|_2^2 .
\label{eq:mse_loss_2}
\end{equation}

% \cref{sec:introduction} \cref{sec:method}
As discussed in Sec.1, the random sampling strategy introduces uncertainty in reconstruction since one single conditional frame can lead to multiple potential predictions. Besides, previous MVM methods primarily restore the masked regions at the pixel level, making the model retain excessive low-level information. To address these issues, as shown in \cref{fig:compare_methods_3}, we propose a framework named T-CoRe with two key properties: \textbf{1)} a sandwich sampling strategy to establish temporal correspondence from auxiliary frames, thereby reducing the reconstruction uncertainty; \textbf{2)} an auxiliary branch on top of a self-distillation structure to reconstruct the masked patches in the latent space, facilitating the capture high-level semantic representations.
Following the detailed illustration in Sec.3, we employ CE loss for representation reconstruction and MSE loss for temporal squeezing:

\begin{equation}
    \begin{aligned}
        \lpt &= -\frac{1}{|\mathcal M(\bm v_c)|}\sum_{i \in \mathcal{M}(\bm v_c)} (\hat {\bm p}_c)^i \log(\bm p_c^p)^i  , \\
        \lft &= -\frac{1}{|\mathcal M(\bm v_c)|}\sum_{i \in \mathcal{M}(\bm v_c)} (\hat {\bm p}_c)^i \log (\bm p_c^f)^i ,
    \label{eq:loss1}
    \end{aligned}
\end{equation}
\begin{equation}
    \lpf = \frac{1}{|\mathcal M(\bm v_c)|}\sum_{i \in \mathcal M(\bm v_c)} \| (\bm p_c^p)^i - (\bm p_c^f)^i \|_2^2 ,
\label{eq:loss2}
\end{equation}
where $\bm p_c^p$, $\bm p_c^f$ are the restored representations of the current frame guided by the past and future frame, respectively, and $\hat {\bm p}_c$ is the representation from the teacher branch.

\section{Detailed Description of Experiments}
\subsection{Training Datasets}
\textbf{Kinetics-400}~\cite{Kinetics} is a large-scale video dataset with 400 categories of daily actions, widely used for tasks like action recognition and video understanding. It includes 239,789 available training videos, each about 10 seconds long. We extract frames at $FPS=2$ to create our training set.

\textbf{ImageNet-1k}~\cite{Imagenet} is a widely used large-scale static image dataset containing 1,000 categories, covering a broad range of real-world objects. The training subset consists of 1.28 million images in the training subset, which we use to pre-train our framework.

\subsection{Evaluation Datasets and Metrics}

\textbf{DAVIS-2017}~\cite{DAVIS17} is a benchmark dataset for video object segmentation. The following three metrics are commonly used to evaluate the overall segmentation performance.

\noindent\textbf{1)} $\mathcal{J}_{\mathrm{m}}$ measures the average region similarity by calculating the overlap between the predicted segmentation mask $P_i$ and the ground truth mask $G_i$ for each video $\bm{V}_i$:
\begin{equation}
    \mathcal{J}_{\mathrm{m}} = \frac{1}{n} \sum_{i=1}^{n} \frac{|P_i \cap G_i|}{|P_i \cup G_i|} .
\label{eq:J_mean}
\end{equation}

\noindent\textbf{2)} $\mathcal{F}_{\mathrm{m}}$ accesses the average contour accuracy by calculating the harmonic mean of the precision $Pre_i$ and recall $Rec_i$ for the predicted boundary of each video $\bm{V}_i$:
\begin{equation}
    \mathcal{F}_{\mathrm{m}} = \frac{1}{n} \sum_{i=1}^{n} \frac{2 \cdot Pre_i \cdot Rec_i}{Pre_i + Rec_i} .
\label{eq:F_mean}
\end{equation}

\noindent\textbf{3)} $\mathcal{J}\&\mathcal{F}_{\mathrm{m}}$ provides a comprehensive measure by averaging $\mathcal{J}_{\mathrm{m}}$ and $\mathcal{F}_{\mathrm{m}}$:
\begin{equation}
    \mathcal{J}\&\mathcal{F}_{\mathrm{m}} = \frac{\mathcal{J}_{\mathrm{m}} + \mathcal{F}_{\mathrm{m}}}{2} .
\label{eq:JF_mean}
\end{equation}

\textbf{JHMDB}~\cite{JHMDB} is primarily used for action recognition and human pose estimation. In our work, we use it for the human pose propagation task, which is evaluated using the PCK@$k$ metric to measure the precision of predictions:
\begin{equation}
    \text{PCK@}k = \frac{1}{n} \sum_{i=1}^{n} \frac{1}{|S_i|} \sum_{j=1}^{|S_i|} \mathbbm{1} \left[ D(\hat{p}_{i,j}, p_{i,j}) < k \cdot d_i \right],
\label{eq:PCK}
\end{equation}
where $S_i$ is the set of key points and $d_i$ is the scale of the human body in video $\bm{V}_i$, $D(\hat{p}_{i,j}, p_{i,j})$ represents Euclidean distance between predicted key point $\hat{p}_{i,j}$ and ground truth key point $p_{i,j}$. The parameter $k$ is the threshold for the maximum allowable distance error.
Following previous works~\cite{SiamMAE,CropMAE,RSP}, we use PCK@0.1 and PCK@0.2 as evaluation metrics.

% =\sqrt{(x_{i,j}-x_{i,j}')^2+(y_{i,j}-y_{i,j}')^2}

\textbf{VIP}~\cite{VIP} is designed for fine-grained instance parsing, which can be applied to the semantic part propagation task. The primary evaluation metric is mIoU, which measures the segmentation performance by calculating the overlap between the predicted segmentation mask $P_{i,j}$ and the ground truth mask $G_{i,j}$ for each video $\bm{V}_i$ and each class $C_j$:
\begin{equation}
    \text{mIoU} = \frac{1}{|C|} \sum_{j=1}^{|C|} \frac{1}{n} \sum_{i=1}^{n} \frac{|P_{i,j} \cap G_{i,j}|}{|P_{i,j} \cup G_{i,j}|} .
\label{eq:mIoU}
\end{equation}

\subsection{Training Settings}
\label{subsec:training}

\noindent\textbf{Sampling strategy.} 
For each video, we first randomly select a current frame within the range $[0.3, 0.7]$ of the total video duration. Then, we randomly sample the past and future frames relative to the current frame based on the offset range of $[0.15, 0.25]$. Moreover, we generate 2 global views ($224\times 224$) and 8 local views ($96\times 96$) from the current frame, applying standard data augmentations such as color jittering, gray scaling, and Gaussian blur.
Following ~\cite{DINOv2}, $50\%$ of the global crops in the student branch are masked with $10\%$ to $50\%$ of randomly selected patches.

\noindent\textbf{Optimizing settings.} 
We adopt ViT-Small and ViT-Base~\cite{Vision_Transformer} with a patch size of 16 as the backbone models in our framework. The feature dimensions are set to 384 and 768, respectively.
For the ViT-S/16 backbone, we pre-train our framework for 400 epochs with a batch size of $256$, where the first 20 epochs are allocated for warm-up. For the ViT-B/16 backbone, we pre-train our framework for 200 epochs with a batch size of $128$, with the first 10 epochs used for warm-up.
The base learning rate $blr$ for ViT-S/16 and ViT-B/16 are adaptively set to $2\times 10^{-3}$ and $1\times 10^{-3}$, respectively, and decay to $1\times 10^{-6}$ using a cosine decay schedule. The real learning rate $lr$ is scaled according to the batch size: $lr=blr \cdot \sqrt{bs/1024}$. The learning rates for PMM are set to $0.1 \times lr$ for ViT-S/16 and $0.13 \times lr$ for ViT-B/16.
The student branch is optimized by AdamW~\cite{AdamW} and the teacher branch is updated with the exponential moving average of the student weights.

\noindent\textbf{Loss function.} 
The hyper-parameters of the loss function are set as follows: $\lambda_1 = 0.8$, $\lambda_2 = 20$, $\lambda_3 = 1.0$, and $\lambda_4 = 0.1$. Note that $\lambda_3$ and $\lambda_4$ are taken from the default setting of \cite{DINOv2} and we only tune the $\lambda_1$ and $\lambda_2$ in our experiments.

\begin{table}[htbp]
  \centering
  \setlength{\tabcolsep}{3pt}
  \resizebox{0.44\textwidth}{!}{%
    \begin{tabular}{lc|cc}
    \toprule
    \multicolumn{1}{c}{\multirow{2}[4]{*}{Hyperparameter}} & \multicolumn{1}{c|}{\multirow{2}[4]{*}{Notation}} & \multicolumn{2}{c}{Value} \\
\cmidrule{3-4}          &       & ViT-S/16 & ViT-B/16 \\
    \midrule
    \rowcolor[rgb]{ .867,  .922,  .969} \multicolumn{4}{c}{Sampling strategy} \\
    \midrule
    Current frame range &    /   & \multicolumn{2}{c}{[0.3, 0.7]} \\
    Past frame offset range &   $[\alpha, \beta]$    & \multicolumn{2}{c}{$[0.15, 0.25]$} \\
    Future frame offset range &   $[\alpha, \beta]$    & \multicolumn{2}{c}{$[0.15, 0.25]$} \\
    Mask probability &    /   & \multicolumn{2}{c}{0.5} \\
    Mask ratio &    /   & \multicolumn{2}{c}{$[0.1, 0.5]$} \\
    Global crop size &   /    & \multicolumn{2}{c}{$(224 \times 224)$} \\
    Local crop size &   /    & \multicolumn{2}{c}{$(96 \times 96)$} \\
    Past and future frame size &   /    & \multicolumn{2}{c}{$(224 \times 224)$} \\
    \midrule
    \rowcolor[rgb]{ .867,  .922,  .969} \multicolumn{4}{c}{Optimizing settings} \\
    \midrule
    Optimizer &   /    & \multicolumn{2}{c}{AdamW} \\
    Learning rate scheduler &    /   & \multicolumn{2}{c}{Cosine} \\
    Weight decay &   /    & \multicolumn{2}{c}{$0.04 \rightarrow 0.4$} \\
    Momentum &   /    & \multicolumn{2}{c}{$0.992 \rightarrow 1$} \\
    Number of ViT encoder blocks &   /    & \multicolumn{2}{c}{12} \\
    Patch size &   $p$    & \multicolumn{2}{c}{16} \\
    Base learning rate  &   $blr$    & $2 \times 10^{-3}$  & $1 \times 10^{-3}$ \\
    PMM learning rate &   /    & $0.1 \times lr$   & $0.13 \times lr$  \\
    Epochs &   /    & 400   & 200 \\
    Warm-up epochs &   /    & 20    & 10 \\
    Batch size &   bs   & 256   & 128 \\
    Number of ViT feature dim. &   $d$    & 384   & 768 \\
    \midrule
    \rowcolor[rgb]{ .867,  .922,  .969} \multicolumn{4}{c}{Loss function} \\
    \midrule
    Weight of reconstruction loss &   $\lambda_1$    & \multicolumn{2}{c}{0.8} \\
    Weight of squeezing loss &   $\lambda_2$     & \multicolumn{2}{c}{20} \\
    Weight of DINO loss &    $\lambda_3$    & \multicolumn{2}{c}{1} \\
    Weight of koleo loss &    $\lambda_4$    & \multicolumn{2}{c}{0.1} \\
    \bottomrule
    \end{tabular}%
  }
  \caption{The hyperparameters settings for our T-CoRe framework during the training process.}
  \label{tab:training_settings}%
  \vspace{-4pt}
\end{table}%

\begin{table}[htbp]
  \centering
  \resizebox{0.4\textwidth}{!}{%
    \begin{tabular}{c|ccc}
    \toprule
    Config & DAVIS-2017 & VIP & JHMDB \\ 
    \midrule
    Top-K & 7 & 10 & 7 \\
    Queue Length & 20 & 20 & 20 \\
    Neighborhood Size & 20 & 20 & 20 \\
    \bottomrule
    \end{tabular}%
    }
    \caption{The hyperparameters settings for our T-CoRe framework during the evaluation process.}
  \label{tab:evaluation_setting}%
  \vspace{-10pt}
\end{table}%

The general hyperparameters settings for our T-CoRe framework during the training process are summarized in ~\cref{tab:training_settings}. All experiments in this work are conducted with Pytorch~\cite{pytorch} on a Linux machine equipped with an AMD EPYC 9654 96-Core Processor and 4 NVIDIA 4090 GPUs.

\subsection{Evaluation settings}

Following~\cite{CropMAE}, the hyperparameters for three downstream tasks are listed in ~\cref{tab:evaluation_setting}.
Note that these hyperparameters remain fixed in our framework without further tuning to ensure a fair comparison.

\subsection{Competitors}
We compare our T-CoRe with various state-of-the-art self-supervised representation learning methods, which can be categorized into the following two types: 

\noindent\textbf{1) Contrastive learning methods}: 

\begin{itemize}
  \item SimCLR~\cite{SimCLRv1} aims to learn meaningful representations by contrasting positive and negative samples.
  \item MoCo v3~\cite{MoCov3} designs a momentum encoder with a memory bank to store negative samples.
  \item DINO~\cite{DINO} uses a self-distillation structure to learn representations with Vision Transformer~\cite{Vision_Transformer} as the encoder.
  \item ODIN$^\text{2}$~\cite{ODIN2} combines object discovery with representation networks to capture meaningful semantics without annotation.
  \item CrOC~\cite{Croc} proposes a cross-view consistency objective with an online clustering mechanism for semantic segmentation.
\end{itemize}

\noindent\textbf{2) Masked modeling methods}:

\begin{itemize}
  \item MAE~\cite{MAE} proposes to mask and recover image patches at the pixel level based on an encoder-decoder structure.
  \item MAE-ST~\cite{MAE_ST} extends MAE to learn spatiotemporal representations from videos.
  \item RC-MAE~\cite{RC-MAE} introduces a mean teacher network into MAE for consistent reconstruction.
  \item VideoMAE~\cite{VideoMAE} simply extends MAE into the video domain by masking video tubes with an extremely high masking ratio and recovering the masked pixels.
  \item DropMAE~\cite{DropMAE} applies adaptive spatial-attention dropout to enhance temporal relations in videos
  \item SiamMAE~\cite{SiamMAE} uses a past frame and a masked current frame as input to a Siamese network, reconstructing the masked patches with a conditional decoder. 
  \item CropMAE~\cite{CropMAE} samples different crops or augmented versions of a frame as input to a similar structure with SiamMAE.
  \item RSP~\cite{RSP} learns to recover a future frame through stochastic frame prediction, using the current frame for prior and posterior distributions.
  \item iBOT~\cite{iBOT} aligns both cross-view \texttt{[CLS]} tokens and in-view patch tokens within a self-distillation framework.
  \item DINO v2~\cite{DINOv2} employs a discriminative self-supervised pre-training approach and incorporates additional techniques~\cite{centering,SwAV,KoLeo,adapting_resolution} to improve iBOT. For a fair comparison, we do not apply the costly techniques in DINO v2 like distillation from a larger model or pre-training with a larger resolution. 
\end{itemize}

\section{Additional Experimental Results}

\begin{figure*}
  \centering
    \begin{subfigure}{0.33\linewidth}
    \includegraphics[width=1.0\linewidth]{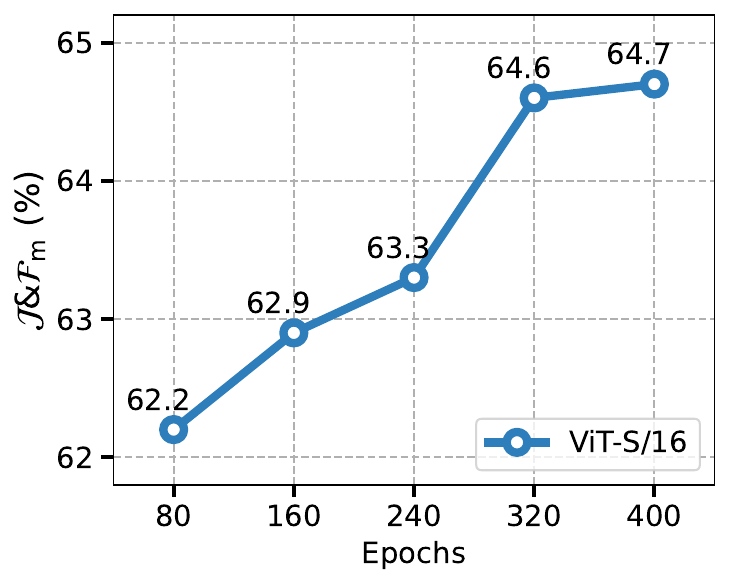}
    \caption{$\mathcal{J}\&\mathcal{F}_{\mathrm{m}}$ on DAVIS-2017 with ViT-S/16.}
    \label{fig:plot_DAVIS_S}
    \end{subfigure}
  % \hfill
  \begin{subfigure}{0.33\linewidth}
    \includegraphics[width=1.0\linewidth]{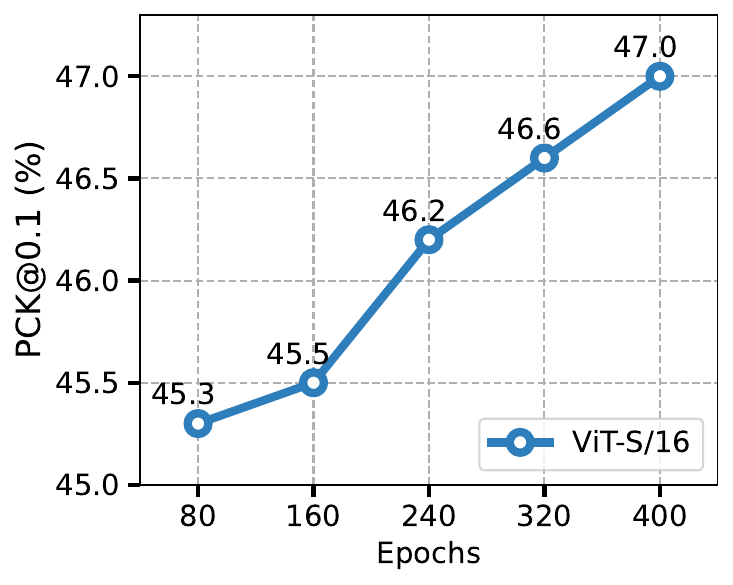}
    \caption{PCK@0.1 on JHMDB with ViT-S/16.}
    \label{fig:plot_JHMDB_S}
  \end{subfigure}
  % \hfill
  \begin{subfigure}{0.33\linewidth}
    \includegraphics[width=1.0\linewidth]{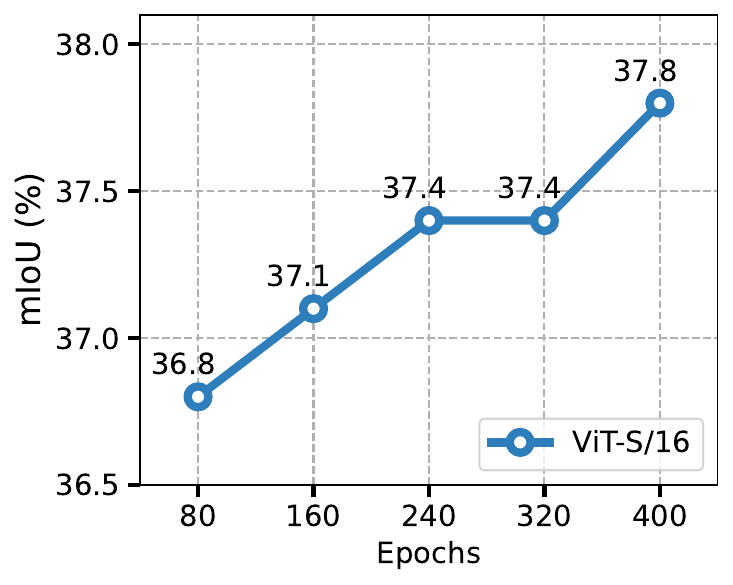}
    \caption{mIoU on VIP with ViT-S/16.}
    \label{fig:plot_VIP_S}
  \end{subfigure}
  
    \par
    
  \begin{subfigure}{0.33\linewidth}
    \includegraphics[width=1.0\linewidth]{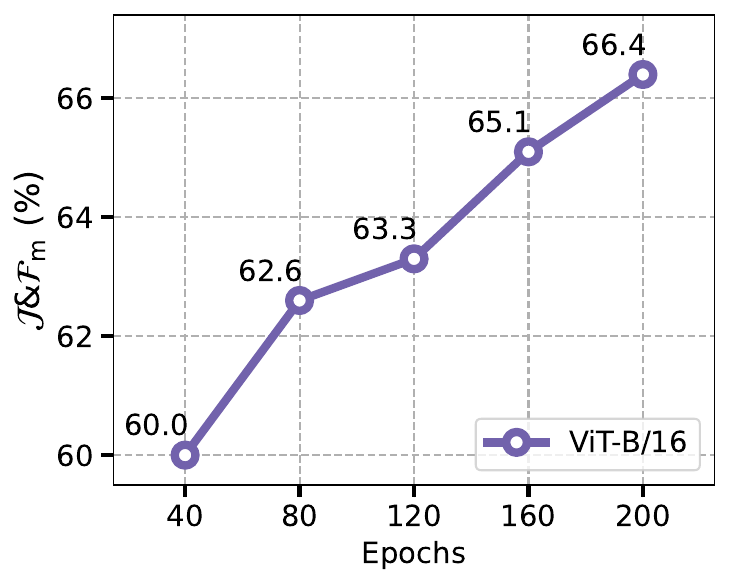}
    \caption{$\mathcal{J}\&\mathcal{F}_{\mathrm{m}}$ on DAVIS-2017 with ViT-B/16.}
    \label{fig:plot_DAVIS_B}
    \end{subfigure}
  % \hfill
  \begin{subfigure}{0.33\linewidth}
    \includegraphics[width=1.0\linewidth]{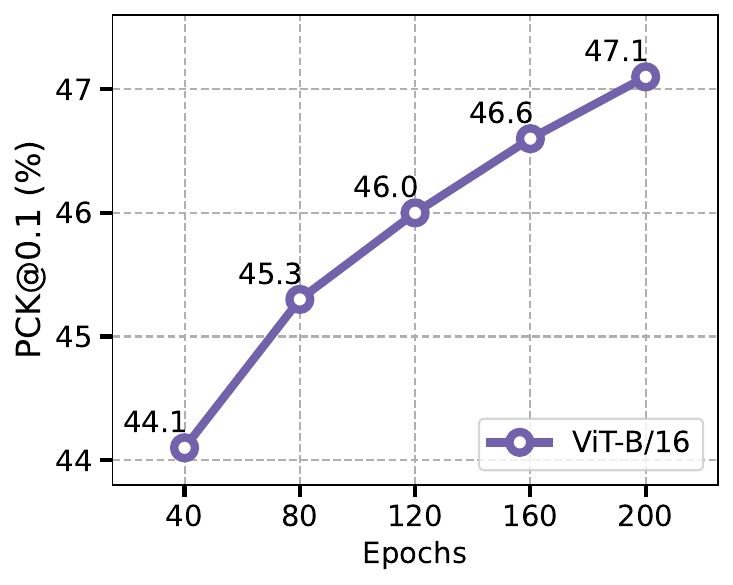}
    \caption{PCK@0.1 on JHMDB with ViT-B/16.}
    \label{fig:plot_JHMDB_B}
  \end{subfigure}
  % \hfill
  \begin{subfigure}{0.33\linewidth}
    \includegraphics[width=1.0\linewidth]{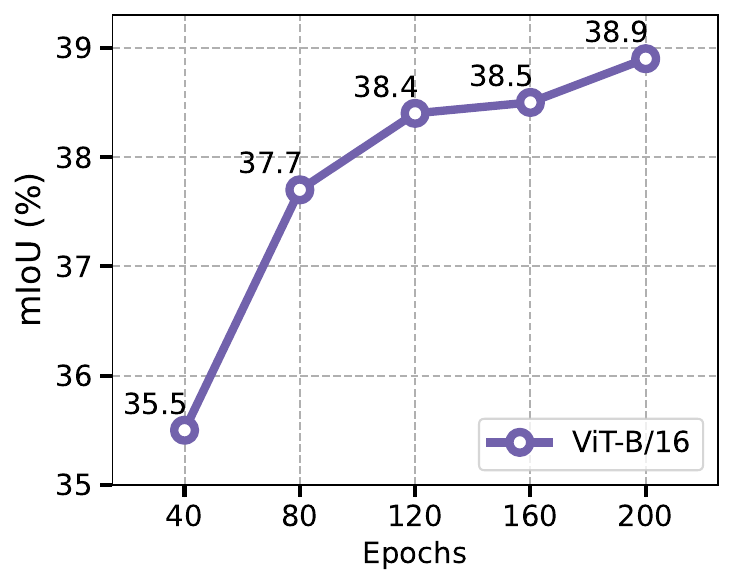}
    \caption{mIoU on VIP with ViT-B/16.}
    \label{fig:plot_VIP_B}
  \end{subfigure}
  
  \caption{The performance on three downstream tasks during the training phase with ViT-S/16 and ViT-B/16 backbones.}
  \vspace{-4pt}
  \label{fig:dynamic}
\end{figure*}

\subsection{Training Dynamics}

\cref{fig:dynamic} illustrates the performance dynamics of the training schedule for ViT-S/16 and ViT-B/16 backbones across three downstream tasks. We report the performance of the framework at different checkpoints during the training process. The figures indicate that training with a larger model and a longer duration leads to further performance improvements on these downstream tasks.

\subsection{Computational Efficiency}

Our framework (VIT-B/16 backbone) is pre-trained on 4 GPUs for 200 epochs with $bs$=128 in 30 hours.
As shown in the \cref{tab:supp_results_efficiency}, we provide a comparison of model efficiency. Although the self-distillation architecture introduces a slight computational overhead, our method achieves superior performance on downstream tasks with fewer training epochs, while maintaining relatively acceptable time and space costs, thus seeking a balance between computational efficiency and model effectiveness.

\begin{table}[htbp]
\vspace{6pt}
  \centering
  \setlength{\tabcolsep}{4pt}
  \resizebox{0.48\textwidth}{!}{%
    \begin{tabular}{cc|ccc|ccc}
    \toprule
        Method  & Backbone & GPU Mem. & Epoch & PT-Time  & $\mathcal{J}\&\mathcal{F}_{\mathrm{m}}$ & $\mathcal{J}_{\mathrm{m}}$   & $\mathcal{F}_{\mathrm{m}}$ \\
          \midrule
           VideoMAE  & ViT-B/16 & 8$\times$24.39 GB  &  800 &  160 h  &  34.7 & 33.9  & 35.4 \\
           SiamMAE   & ViT-B/16 & 4$\times$5.04 GB  &  400 &  21 h  &  45.5 & 43.6 & 47.5 \\
           CropMAE  & ViT-B/16 & 4$\times$5.02 GB  &  400 &  22 h &  57.8 & 56.9 & 58.7 \\
           RSP   & ViT-B/16 & 4$\times$12.19 GB  &  400 &  113 h &  \underline{60.5} & \underline{57.8} & \underline{63.2} \\
           \textbf{T-CoRe (Ours)}  & ViT-B/16 & 4$\times$17.57 GB  &  200 & 30 h &  \textbf{66.4} & \textbf{64.6} & \textbf{68.2} \\
    \bottomrule
    \end{tabular}%
    }
    \vspace{-2pt}
    \caption{Efficiency comparison on ViT-B/16 with same settings.}
  \label{tab:supp_results_efficiency}%
  \vspace{-4pt}
\end{table}%

\subsection{Additional Visualization Results}

\begin{figure*}
% \vspace{-10pt}
  \centering
    \includegraphics[width=0.94\linewidth]{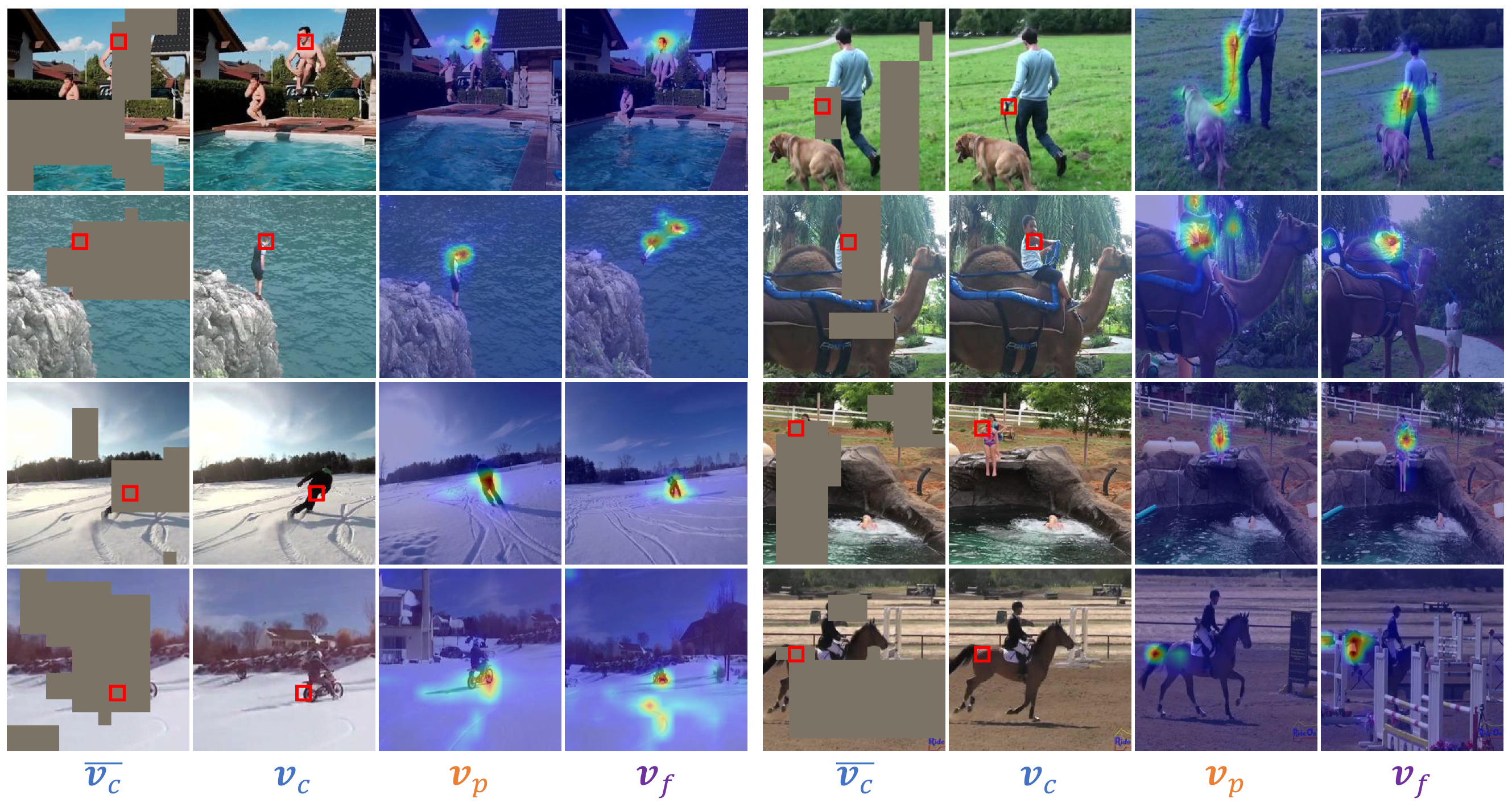}
     \vspace{-4pt}
    \caption{Additional cross-attention heatmaps of the masked current frame $\bar{\bm v}_c$ to the past and future frames $\vp, \vf$ in the PMM.}
    \label{fig:vis_heatmaps_supp}
\end{figure*}

\textbf{Cross-attention Maps.}
We provide additional cross-attention heatmaps of the masked patches between the current frame and both the past and future frames in PMM.
As shown in \cref{fig:vis_heatmaps_supp}, the masked patches could successfully match similar regions in the auxiliary frames through the cross-attention mechanism, demonstrating the favorable ability to establish temporal correspondence. This capability remains effective in perceiving and matching the corresponding targets even when the target is almost completely masked in the current frame. Moreover, it is worth noting that PMM can also capture edge details, such as the outline of the rope in the top-right example.

\begin{figure*}
  \centering
    \includegraphics[width=0.94\linewidth]{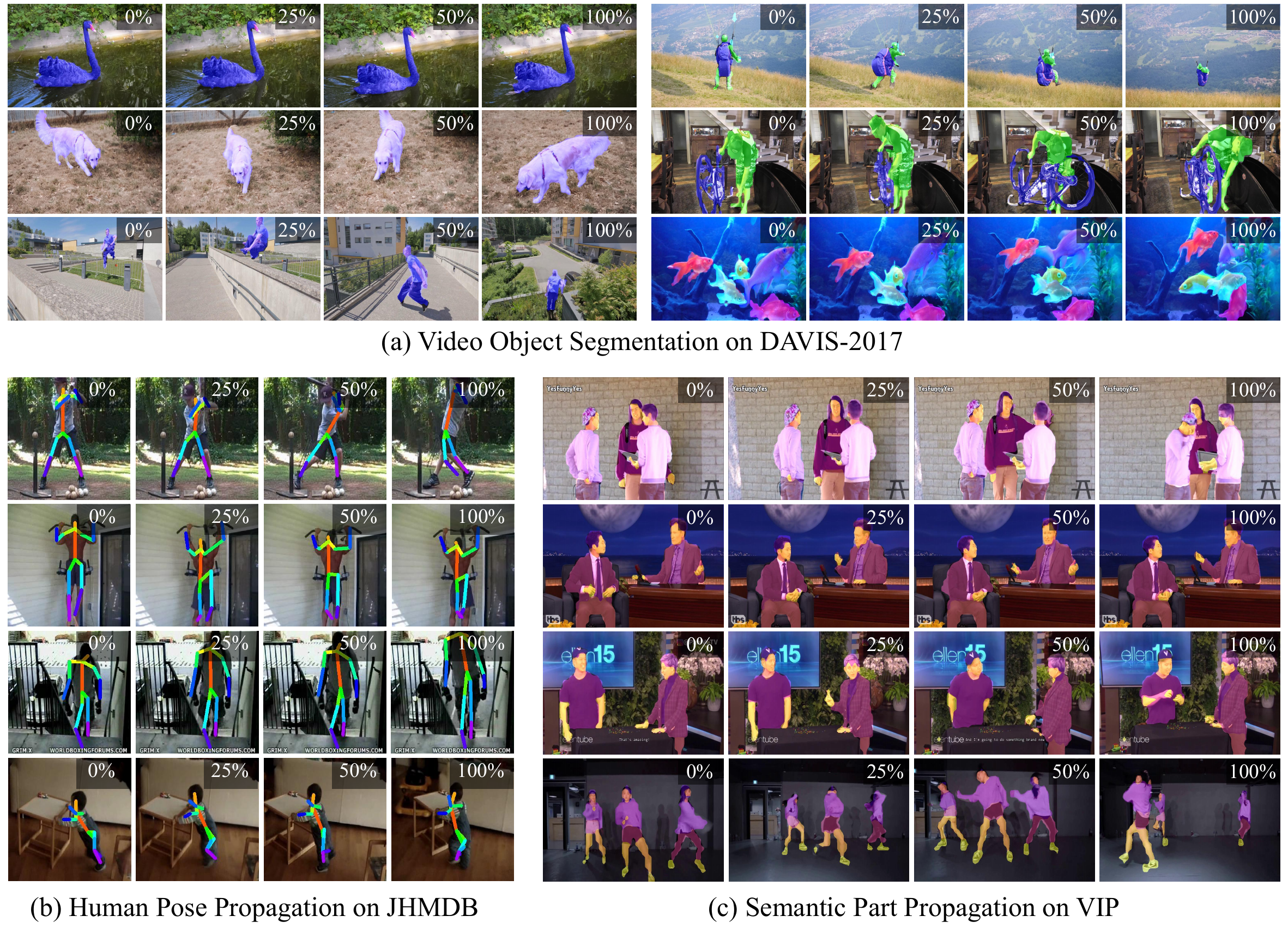}
    \vspace{-4pt}
    \caption{Additional visualization results of T-CoRe for three downstream tasks including (a) video object segmentation on DAVIS-2017~\cite{DAVIS17}, (b) human pose propagation on JHMDB~\cite{JHMDB}, and (c) body part propagation on VIP~\cite{VIP}.}
    \label{fig:vis_results_supp}
    % \vspace{-10pt}
\end{figure*}

\textbf{Downstream Tasks.}
In \cref{fig:vis_results_supp}, we provide more visualization results on three downstream tasks. The prediction masks show that T-CoRe performs well in instance segmentation and posture tracking across most scenarios, making it an effective pre-training framework for video representation learning that facilitates the video understanding process.

\subsection{$k$-NN Images for Pre-training on ImageNet-1k}

\begin{figure*}
% \vspace{-10pt}
  \centering
    \includegraphics[width=0.86\linewidth]{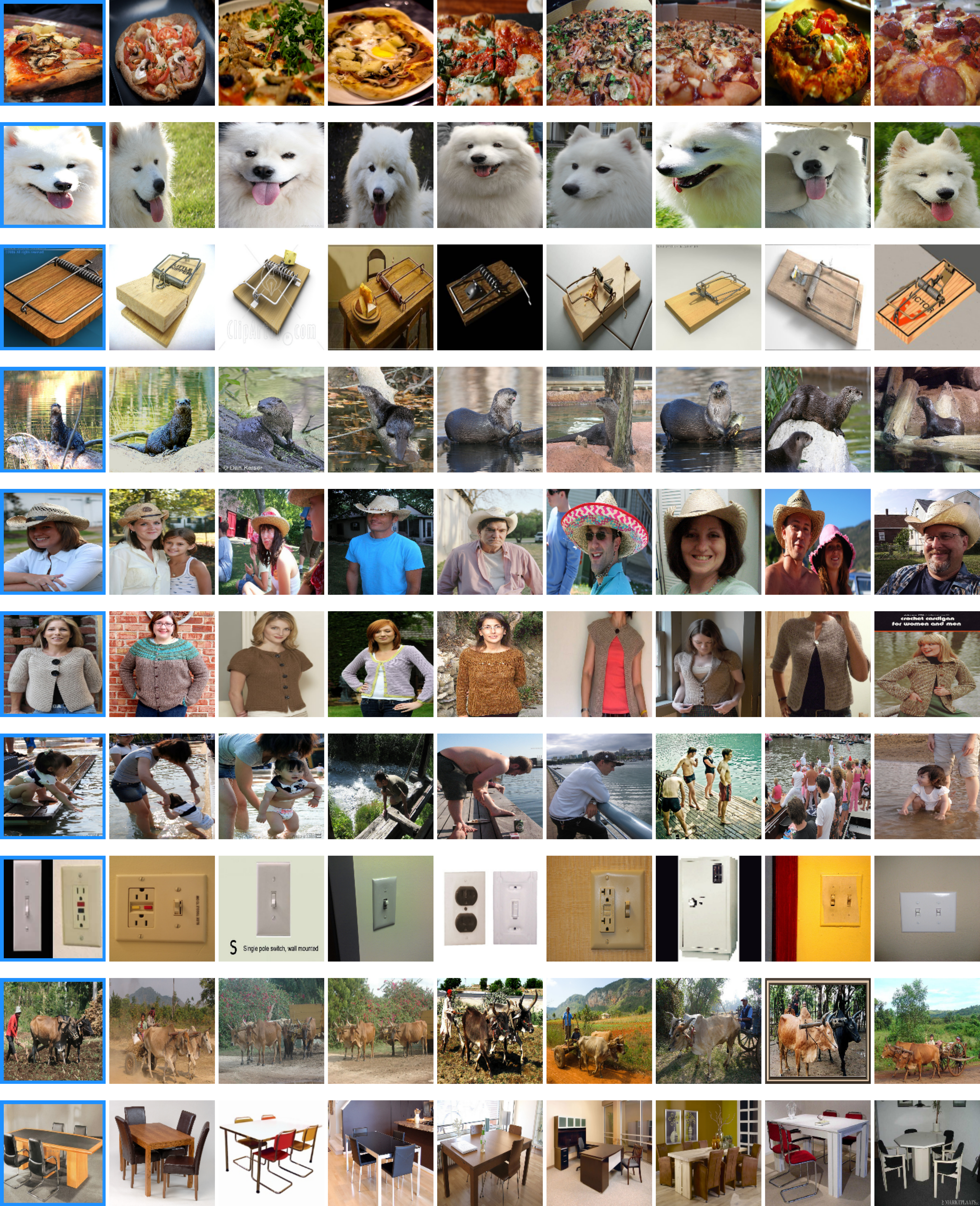}
    \vspace{-2pt}
    \caption{Examples of $k$-NN images for pre-training on ImageNet-1k. The images within the blue boxes in the first column are the origin images, while the following 8 columns show their top-$k$ nearest neighbor images in the training set.}
    \label{fig:vis_knn}
\end{figure*}

To ensure a fair comparison with previous methods in the image domain, we extend our framework for pre-training on the ImageNet-1k~\cite{Imagenet}.
Specifically, we employ $k$-NN images to simulate the adjacent frames in videos for establishing temporal correspondence. In this setting, the auxiliary branch receives one $k$-NN image. The $k$-NN images are determined based on our framework pre-trained without the auxiliary branch, where $k$ is set to 5 at default. 
\cref{fig:vis_knn} presents several examples of original images and their corresponding $k$-NN images, which exhibit similar appearance and share the same semantics as the original images, effectively simulating the selection of adjacent frames from a video to establish correspondence.

\begin{figure*}
\vspace{-4pt}
  \centering
    \includegraphics[width=0.96\linewidth]{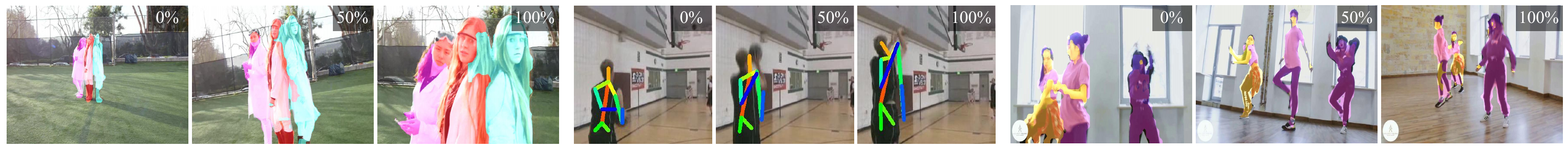}
    \vspace{-4pt}
    \caption{Failure cases on downstream tasks.}
    \label{fig:failure}
\end{figure*}

\subsection{Failure Case Analysis}

As shown in \cref{fig:failure}, obvious prediction errors occur in some challenging test samples, such as tightly fitting instances, considerable motion amplitudes, and significant camera movements. These failures likely result from the limitations in the training data processing, which prevents the model from learning such difficult scenarios. This issue could be mitigated by using higher-quality datasets, more precise sampling methods, and smaller patch sizes for training. In future work, we will consider further enhancing the framework to better handle more complex scenarios.

\end{document}